\documentclass[11pt]{article}

\usepackage[preprint]{acl}

\usepackage{times}
\usepackage{latexsym}
\usepackage[T1]{fontenc}
\usepackage[utf8]{inputenc}
\usepackage{microtype}
\usepackage{inconsolata}
\usepackage{graphicx}

\usepackage{amsmath}
\usepackage{amssymb}
\usepackage{booktabs}
\usepackage{multirow}
\usepackage{tabularx}
\usepackage{array}
\usepackage{pifont}
\usepackage[most]{tcolorbox}

\newcommand{\cmark}{\ensuremath{\checkmark}}
\newcommand{\xmark}{\ensuremath{\times}}

\newcommand{\MthreeRBench}{$\text{M}^3\text{R}$-Bench}

\newcommand{\MthreeRReasoner}{$\text{M}^3\text{R}$-Reasoner}

\colorlet{RevisionRed}{black}

\newtcolorbox{promptbox}[1][]{
    enhanced,
    colback=white,
    colframe=black!60,
    colbacktitle=black!60,
    coltitle=white,
    fonttitle=\bfseries,
    boxrule=0.9pt,
    arc=2mm,
    outer arc=2mm,
    left=7pt,
    right=7pt,
    top=6pt,
    bottom=6pt,
    width=\linewidth,
    #1
}

\providecommand{\cmark}{\ding{51}}
\providecommand{\xmark}{\ding{55}}

\title{M³R-Bench: A Unified Benchmark for Evidence-Grounded Multimodal Metaphor Understanding}

\author{
\textbf{Hong Jiang\textsuperscript{1,*}},
\textbf{Junnan Zhu\textsuperscript{2,*}},
\textbf{Jingwang Huang\textsuperscript{1,*}},
\textbf{Xiao Sun\textsuperscript{1}},
\textbf{Yuming Yang\textsuperscript{1}},
\textbf{Jiang Zhong\textsuperscript{1,$\dagger$}},\\
\textbf{Ruirui Chen\textsuperscript{3}},
\textbf{Jingman Shi\textsuperscript{4}},
\textbf{Hao Wu\textsuperscript{4}},
\textbf{Nayu Liu\textsuperscript{5}},
\textbf{Xinyi Jiang\textsuperscript{6}},
\textbf{Kaiwen Wei\textsuperscript{1,$\dagger$}}\\
\textsuperscript{1}School of Computer Science, Chongqing University, China\\
\textsuperscript{2}Institute of Automation, Chinese Academy of Sciences, China\\
\textsuperscript{3}Institute of Advanced Intelligence and Computing, A*STAR, Singapore\\
\textsuperscript{4}Chongqing Medical University, China\\
\textsuperscript{5}Tianjin University, China\\
\textsuperscript{6}School of Computer Science and Engineering, UNSW, Australia\\
\small{\textbf{Correspondence:} \texttt{jiangh@stu.cqu.edu.cn},
\texttt{jiangzhong@cqu.edu.cn}, \texttt{weikaiwen@cqu.edu.cn}}
}

\begin{document}

\colorlet{red}{black}

\maketitle

\begingroup
\renewcommand{\thefootnote}{\fnsymbol{footnote}}
\footnotetext[1]{These authors contributed equally.}
\footnotetext[2]{Corresponding authors.}
\endgroup

\begin{abstract}
Multimodal metaphor understanding requires identifying cross-domain Target--Source mappings and inferring sentiment from visual and textual evidence. However, existing benchmarks largely evaluate these subtasks separately and lack evidence-grounded explanations, limiting their ability to assess coherent cross-modal reasoning. To enable unified and evidence-grounded evaluation, we introduce \textbf{\MthreeRBench{}}, a benchmark of 1,000 human-annotated image--text instances with joint labels for metaphor occurrence, Target--Source mapping, and sentiment, alongside stage-wise explanations for evidence identification, mapping establishment, and sentiment inference. Evaluations reveal that \textit{existing models frequently rely on textual shortcuts and produce Target--Source mappings inconsistent with visual evidence}. To address this evidence--mapping mismatch, we propose \textbf{\MthreeRReasoner{}}, which combines curriculum-based reasoning supervision with task-aware reinforcement learning. Built on an 8B-parameter backbone, \MthreeRReasoner{} achieves 37.90\% Joint Label Accuracy (JLA) metric, outperforming the strongest proprietary baseline by 19.50 percentage points. It further surpasses GPT-5.5 in image-text evidence and sentiment justification scores by 28.45 and 28.11 points, respectively. The dataset and code are available at \url{https://github.com/hongshi4/m3r-bench-artifact}.

\end{abstract}

\section{Introduction}

\begin{figure}[t!]
    \centering
    \includegraphics[width=\columnwidth]{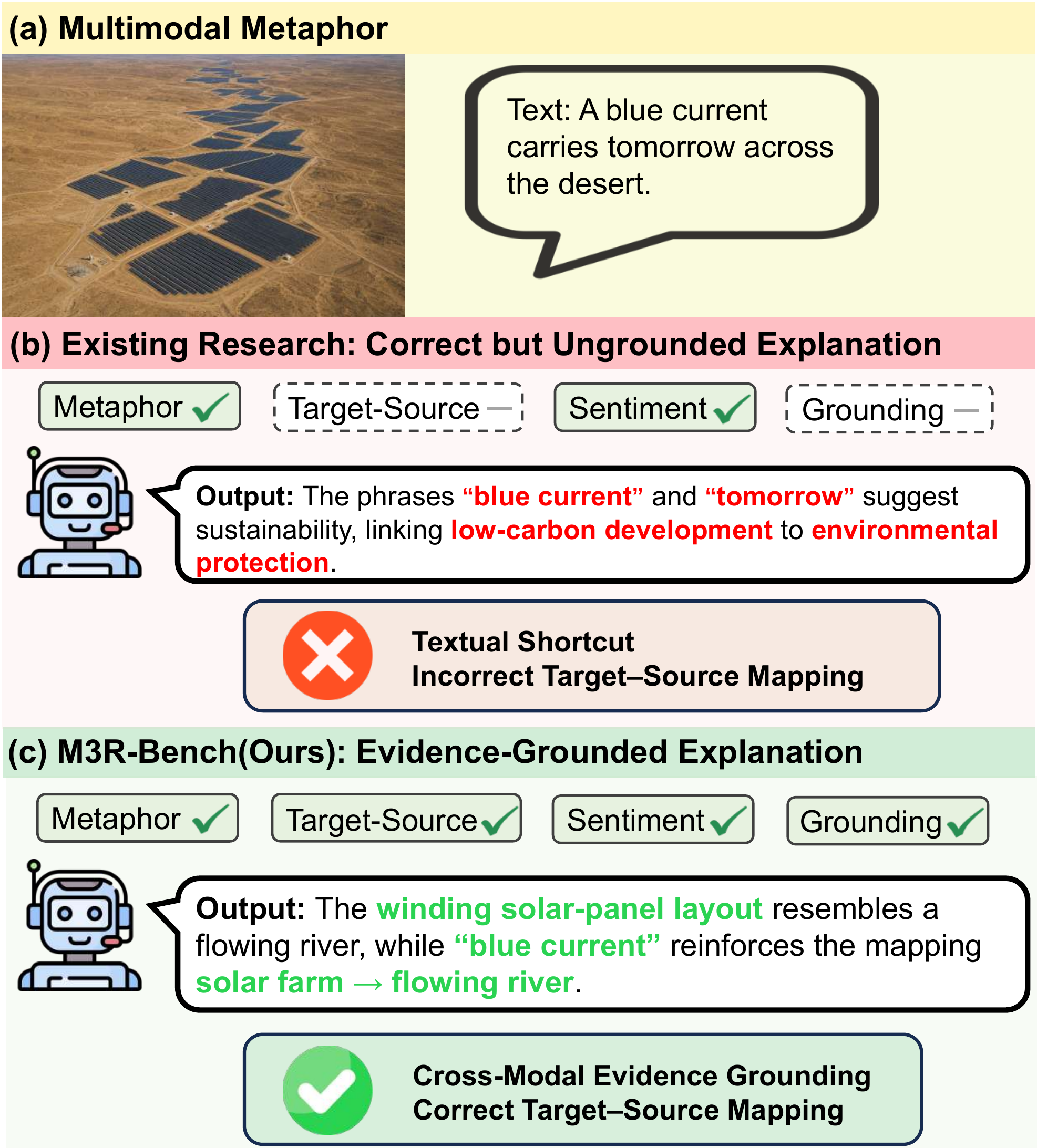}
\caption{Motivation of this work. Existing research predicts metaphor and sentiment but lacks evidence-grounded Target--Source mappings. In contrast, \MthreeRBench{} jointly evaluates metaphor identification, mapping, and sentiment with grounded explanations.
}
    \label{fig:task_overview}
\end{figure}

Metaphor facilitates abstract concept understanding through source-to-target domain mappings \cite{lakoff1980metaphors}. In advertising and social media, such mappings often involve both visual and textual cues, forming multimodal metaphors \cite{forceville2009multimodal}. Understanding these metaphors requires identifying conceptual mappings, affective attitudes, and communicative intentions \cite{zhang2021multimet,xu2022metmeme}.

\begin{table*}[t!]
\centering


\small
\setlength{\tabcolsep}{0.35pt}
\renewcommand{\arraystretch}{0.88}

\begin{tabular*}{\textwidth}{
@{\extracolsep{\fill}}
lllccccc
@{}
}
\toprule
Benchmark
& Modality
& Language
& \shortstack{Metaphor\\Detection}
& \shortstack{Target--Source\\Mapping}
& \shortstack{Sentiment\\/Emotion}
& \shortstack{Unified\\Full-Task}
& \shortstack{Evidence-Grounded\\Rationales} \\
\midrule

TroFi~\cite{birke2006trofi}
& T
& EN
& \cmark
& \xmark
& \xmark
& \xmark
& \xmark \\

VUA All~\cite{steen2010mipvu}
& T
& EN
& \cmark
& \xmark
& \xmark
& \xmark
& \xmark \\

CMDAG~\cite{shao2024cmdag}
& T
& CN
& \xmark
& \cmark
& \xmark
& \xmark
& \xmark \\

MetaCLUE~\cite{akula2023metaclue}
& I
& EN
& \cmark
& \cmark
& \xmark
& \xmark
& \xmark \\

MultiMET~\cite{zhang2021multimet}
& I--T
& EN
& \cmark
& \cmark
& \cmark
& \xmark
& \xmark \\

MET-Meme~\cite{xu2022metmeme}
& I--T
& CN/EN
& \cmark
& \xmark
& \cmark
& \xmark
& \xmark \\

MultiCMET~\cite{zhang2023multicmet}
& I--T
& CN
& \cmark
& \cmark
& \cmark
& \xmark
& \xmark \\

FigMemes~\cite{liu2022figmemes}
& I--T
& EN
& \cmark
& \xmark
& \xmark
& \xmark
& \xmark \\

CM3D~\cite{zhang2025cm3d}
& I--T
& CN
& \xmark
& \cmark
& \xmark
& \xmark
& \xmark \\

CII-Bench~\cite{zhang2025ciibench}
& I--T
& CN
& \cmark
& \xmark
& \cmark
& \xmark
& \xmark \\

MultiMM~\cite{yang2025multimm}
& I--T
& CN/EN
& \cmark
& \cmark
& \cmark
& \xmark
& \xmark \\

EmoMeta~\cite{lu2025emometa}
& I--T
& CN
& \xmark
& \cmark
& \cmark
& \xmark
& \xmark \\

V-FLUTE~\cite{saakyan2025vflute}
&I--T
& EN
& \xmark
& \xmark
& \xmark
& \xmark
& \cmark \\

M3UCD~\cite{zheng2026m3ucd}
& I--T
& CN/EN
& \cmark
& \cmark
& \cmark
& \xmark
& \xmark \\

\midrule

\textbf{\MthreeRBench{} (Ours)}
& I--T
& CN/EN
& \cmark
& \cmark
& \cmark
& \cmark
& \cmark \\

\bottomrule
\end{tabular*}

\caption{Comparison of \MthreeRBench{} with representative multimodal
metaphor and figurative-language benchmarks. I and T denote image and text;
EN and CN denote English and Chinese.}
\label{tab:benchmark_comparison}

\end{table*}

Recent advances in multimodal large language models (MLLMs) have extended metaphor understanding from classification to explicit reasoning \cite{kundu2025imagemet,zheng2026m3ucd}. Existing benchmarks cover metaphor identification, Target--Source mapping, and sentiment analysis \cite{zhang2021multimet,zhang2023multicmet,yang2025multimm}, while recent studies explore expert-verified explanations and staged reasoning \cite{saakyan2025vflute,yang2026multimetu}. However, these tasks are largely evaluated separately, and few benchmarks jointly assess predictions and evidence-grounded, stage-wise explanations. Consequently, component-level performance cannot establish whether models recover coherent mappings from cross-modal evidence. As shown in Figure~\ref{fig:task_overview}, an MLLM may correctly identify a metaphor and its sentiment yet replace the grounded mapping \emph{solar farm} $\rightarrow$ \emph{flowing river} with abstract themes such as \emph{low-carbon development} and \emph{environmental protection}.

To address these limitations, we introduce \textbf{M}ulti\textbf{m}odal \textbf{M}etaphor \textbf{U}nderstanding \textbf{Bench}mark (\textbf{\MthreeRBench{}}), a unified evidence-grounded benchmark. It contains 1,000 image--text instances selected through model disagreement from a deduplicated bilingual candidate pool, disjoint from the 8,989-instance reasoning supervision corpus. Guided by Conceptual Metaphor Theory and Selectional Preference Violation~\cite{lakoff1980metaphors,wilks1975preferential}, each instance provides human-verified annotations for metaphor occurrence, Target--Source mapping, and sentiment, alongside stage-wise explanations following ``evidence identification--mapping establishment--sentiment inference.'' As compared in Table~\ref{tab:benchmark_comparison}, \MthreeRBench{} jointly evaluates these dimensions through component metrics, Joint Label Accuracy (JLA), and evidence-grounded explanation quality, enabling holistic evaluation and fine-grained diagnosis of multimodal metaphor reasoning.

We evaluate 13 baselines spanning vision--language encoders, task-specific systems, and open-weight and proprietary MLLMs on \MthreeRBench{}. Our findings \textit{reveal a persistent cross-modal evidence--mapping mismatch}, where models struggle to establish Target--Source mappings grounded in visual and textual evidence. Specifically, the strongest direct-prompt baseline, GPT-5.5, achieves only 18.40\% JLA, rising to 23.00\% with enhanced prompting. Analysis of 100 GPT-4o errors identifies textual shortcuts (43\%), mapping-granularity mismatch (26\%), and visual-grounding failures (21\%) as dominant failure modes.

To address this mismatch, we propose \textbf{\MthreeRReasoner{}}, which combines curriculum-based reasoning supervision with task-aware reinforcement learning. With a Qwen3-VL-8B~\cite{bai2025qwen3vl} backbone, it reaches 37.90\% JLA, outperforming GPT-5.5 by 19.50 points under the matched direct-prompt setting. Independent human and Gemini scores closely agree on 200 stratified instances, supporting Gemini-based full-benchmark explanation evaluation. \MthreeRReasoner{} achieves a mean score of 56.77, exceeding Claude-Sonnet-4.6 by 7.00 points. These results show that task-specific training improves both joint prediction and stage-wise explanation quality. In summary, the  contributions of this work are:

(1) We identify fragmented evaluation and limited interpretability in existing multimodal metaphor benchmarks, and address these limitations by introducing \textbf{\MthreeRBench{}}, a unified evidence-grounded benchmark with joint annotations for metaphor identification, Target--Source mapping, sentiment, and reasoning explanations.

(2) We evaluate 13 baselines on \MthreeRBench{}, revealing a persistent cross-modal evidence--mapping mismatch characterized by overlooked visual evidence, reliance on superficial textual cues, and inaccurate Target--Source mappings.

(3) We propose \textbf{\MthreeRReasoner{}}, a curriculum-based reasoning framework with task-aware reinforcement learning. Experiments reveal that \MthreeRReasoner{} effectively alleviates cross-modal evidence--mapping mismatch through evidence-grounded reasoning.

\section{Related Work}

\paragraph{Multimodal Metaphor Benchmarks.}

Multimodal metaphor benchmarks have expanded from detection to richer
annotations of conceptual domains, Target--Source relations, sentiment, and
intent. MultiMET~\cite{zhang2021multimet}, MET-Meme~\cite{xu2022metmeme},
and MultiCMET~\cite{zhang2023multicmet} establish representative image--text
settings; MultiMM~\cite{yang2025multimm}, EmoMeta~\cite{lu2025emometa},
CM3D~\cite{zhang2025cm3d}, M3UCD~\cite{zheng2026m3ucd}, and
MultiMETu~\cite{yang2026multimetu} extend cultural, affective, mapping, and
multi-task coverage. MetaCLUE~\cite{akula2023metaclue} and
ImageMet~\cite{kundu2025imagemet} cover localization, captioning, and question
answering, while V-FLUTE~\cite{saakyan2025vflute} studies explainable visual
entailment. Despite this progress, existing benchmarks largely evaluate isolated outputs and metrics, with few providing human-verified explanations aligned with evidence identification, mapping establishment, and sentiment inference. Consequently, component-wise scores cannot establish whether a model completes the full interpretation chain by recovering coherent mappings from cross-modal evidence rather than relying on superficial correlations.
\MthreeRBench{} addresses this gap through joint structured evaluation and
human-verified stage-wise explanations.

\paragraph{Metaphor Understanding Methods.}

Early methods combine image--text features with task-specific semantic
knowledge, including conceptual-domain and sentiment representations
\cite{zhang2021multimet,xu2022metmeme,zhang2023multicmet,yang2025multimm}.
Recent MLLM-based methods introduce explicit reasoning: C4MMD~\cite{xu2024exploring}
uses staged chain-of-thought reasoning; CoC~\cite{zhang2025coc} and
CPMMIM~\cite{zhang2025cm3d} elicit candidate mappings;
ImaRA~\cite{tian2025imara} reasons over imaginative frames and cross-domain
similarities; and MMRGF~\cite{yang2026multimetu} uses tree-structured reasoning
with backtracking. However, most methods address only part of the task or
optimize detection, mapping, and sentiment separately; their reasoning is
usually elicited through prompts or task-specific outputs. \MthreeRReasoner{}
instead learns the evidence--mapping--sentiment chain through curriculum
supervision and task-aware reinforcement learning.

\begin{figure}[t]
\centering
\includegraphics[
width=\columnwidth,
clip
]{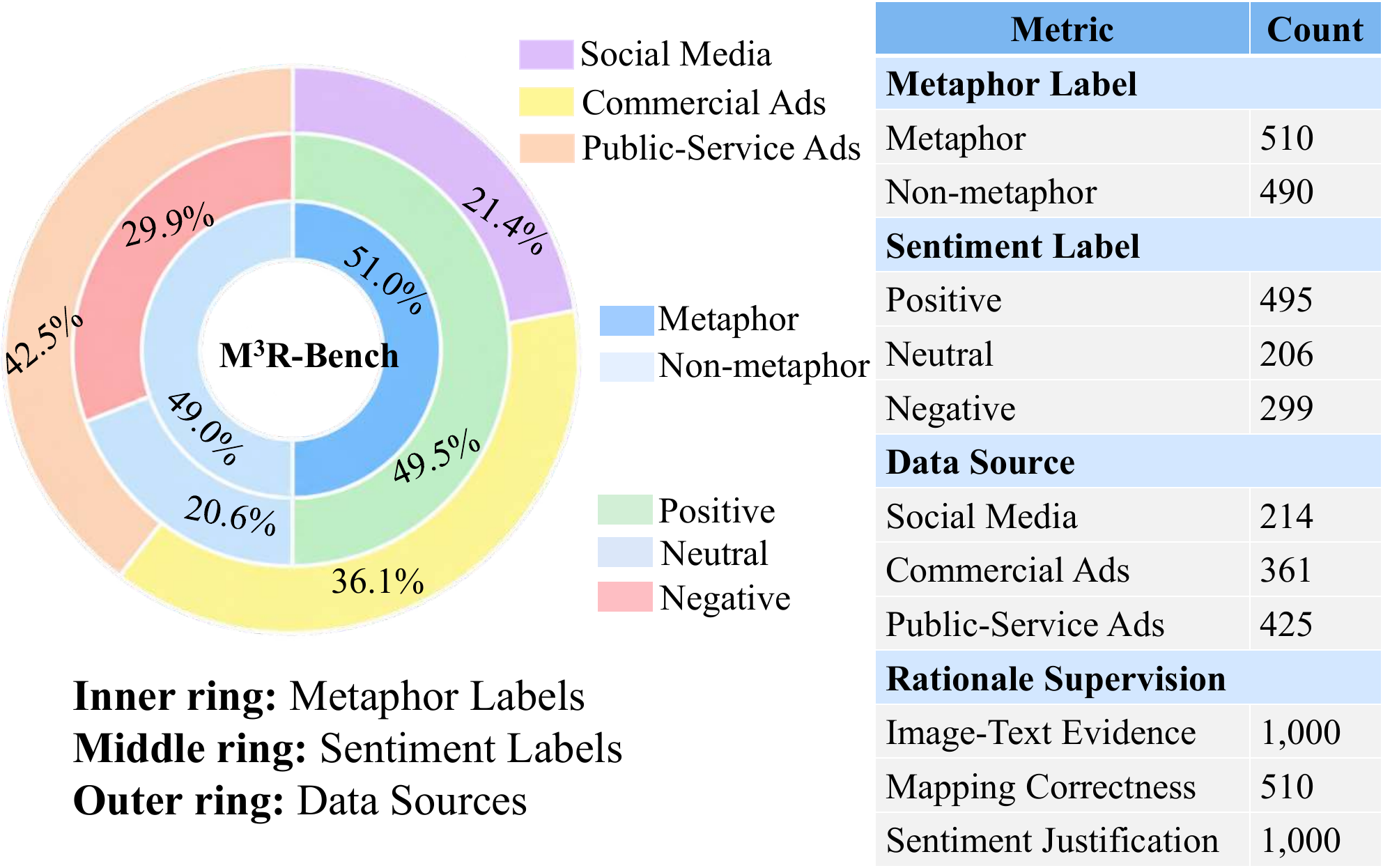}
\caption{Dataset distribution and statistics of \MthreeRBench{} across task labels and source domains.}
\label{fig:m3rbench_overview}
\end{figure}

\begin{figure*}[t]
\centering
\includegraphics[
width=\textwidth,
]{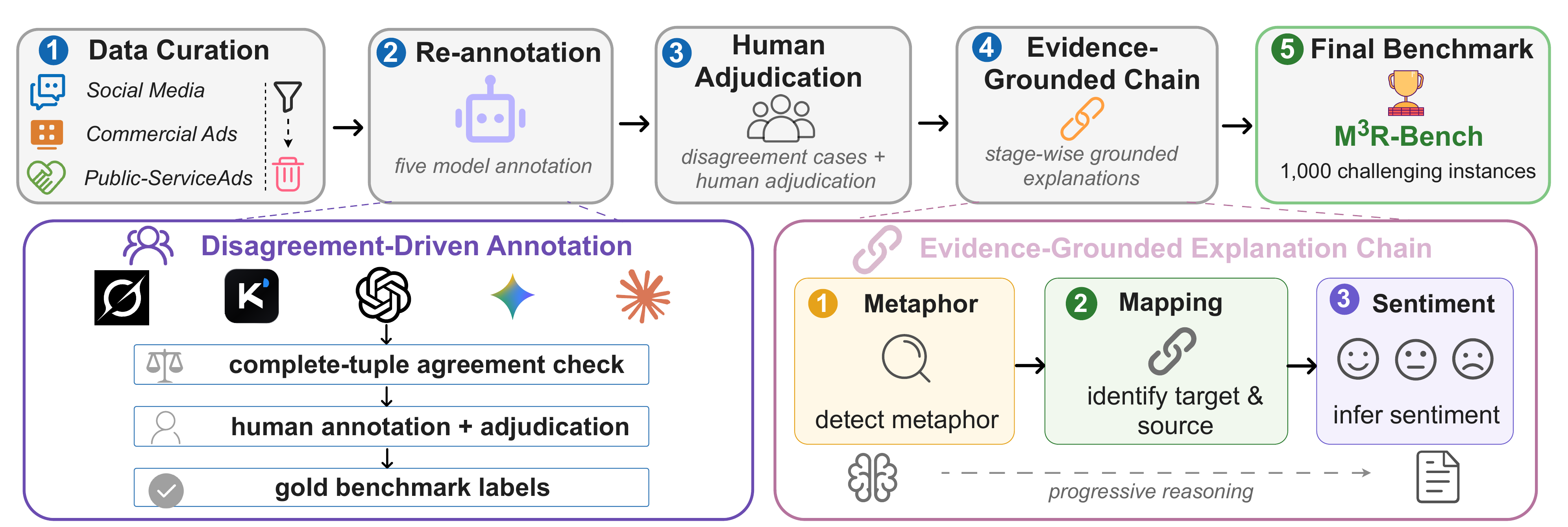}
\caption{Construction pipeline of \MthreeRBench{}.
Curated image--text pairs undergo model-assisted re-annotation,
disagreement-driven human adjudication, and stage-wise evidence-grounded
explanation for metaphor identification, Target--Source mapping, and sentiment
inference.}
\label{fig:bench_construction}
\end{figure*}

\section{\MthreeRBench{}}
\label{sec:m3rbench}


We introduce \MthreeRBench{}, a multimodal metaphor understanding benchmark constructed through model-assisted human annotation and verification. It jointly evaluates metaphor identification, Target--Source mapping, and sentiment inference through unified structured annotations and human-verified explanations for evidence identification, mapping establishment, and sentiment inference.

\paragraph{Data Sources and Curation.}
We consolidate released Chinese and English image--text pairs from
MultiMET~\cite{zhang2021multimet}, MultiCMET~\cite{zhang2023multicmet},
MultiMM~\cite{yang2025multimm}, EmoMeta~\cite{lu2025emometa}, the 2021
iFLYTEK Advertising Image Classification Competition, and the advertising
collection introduced by \citet{ye2021interpreting}. We remove corrupted or
missing images and incomplete pairs, and then perform exact and multimodal
near-duplicate detection over the complete candidate pool, leaving
unique valid pairs. Because the source resources differ in task scope, label
space, and annotation granularity, we discard their inherited labels and
re-annotate every retained sample under a common schema.
Appendix~\ref{app:data} reports the source composition, cleaning rules, and
duplicate-detection protocol.

\paragraph{Model-Assisted Selection and Human Annotation.}
Five MLLMs---Grok-4.1~\cite{xai2025grok41},
Kimi-K2.5~\cite{moonshotai2026kimik25}, GPT-5.5~\cite{openai2026gpt55},
Claude-Opus-4.7~\cite{anthropic2026claudeopus47}, and
Gemini-3.1-Pro~\cite{googledeepmind2026gemini31pro}---independently annotate
each candidate with the complete tuple $(m,t,s,e)$. Five-model agreement
requires the same subset of at least three outputs to agree on every
applicable field: exact agreement for metaphor occurrence and sentiment,
BERTScore F1 $\geq0.75$ separately for Target and Source on metaphors, and
\texttt{None} for both mapping fields on non-metaphors. Candidates that fail
this complete-tuple agreement rule are selected as difficult cases, yielding
the 1,000-instance \MthreeRBench{}.

Three trained annotators independently assign gold labels to the 1,000 benchmark instances under Conceptual Metaphor Theory
(CMT)~\cite{lakoff1980metaphors} and Selectional Preference Violation
(SPV)~\cite{wilks1975preferential}. They view anonymized model proposals and
may reject all of them and write a new complete tuple. Unanimous tuples become
gold labels; a fourth senior annotator adjudicates every remaining case.
Appendix~\ref{app:protocol} gives the complete decision rules, while
Appendix~\ref{app:agreement} reports field-level reliability and the
independent threshold-calibration study.

\paragraph{Human-Verified Gold Explanations.}
GPT-5.4 first predicts the structured tuple together with a three-stage
explanation covering visual and textual evidence, the Target--Source mapping,
and sentiment justification. We automatically compare its predicted tuple
with the finalized human gold labels. When all applicable labels match, two
annotators independently inspect the generated explanation for unsupported or
hallucinated content and make minimal corrections. When any label mismatches,
annotators write the reference explanation directly from the image--text pair
and gold labels. A third annotator adjudicates disagreements in both paths.

\paragraph{Dataset Statistics.}
Figure~\ref{fig:m3rbench_overview} summarizes \MthreeRBench{}: 510
metaphorical and 490 non-metaphorical instances, with 495 positive, 206
neutral, and 299 negative labels. The benchmark covers social media (214),
commercial advertisements (361), and public-service advertisements (425).
Every instance provides gold fields for metaphor occurrence, Target,
Source, and sentiment, together with human-verified Image-Text Evidence and Sentiment
Justification references. For the 490 non-metaphorical instances, Target and
Source are explicitly set to \texttt{None}; the 510 metaphorical instances
additionally provide Mapping Correctness references.

\begin{table*}[t]
\centering

\small
\setlength{\tabcolsep}{2.0pt}

\begin{tabular*}{\textwidth}{
@{\extracolsep{\fill}}
ll
cc
cc
cc
cc
c
c
@{}
}
\toprule
\multirow{2}{*}{Type}
& \multirow{2}{*}{Model}
& \multicolumn{2}{c}{MD}
& \multicolumn{2}{c}{TP}
& \multicolumn{2}{c}{SP}
& \multicolumn{2}{c}{SC}
& \multicolumn{1}{c}{JLA}
& \multicolumn{1}{c}{Rubric} \\
\cmidrule(lr){3-4}
\cmidrule(lr){5-6}
\cmidrule(lr){7-8}
\cmidrule(lr){9-10}
\cmidrule(lr){11-11}
\cmidrule(l){12-12}
&
& Acc. & Macro-F1
& BERT-P & BERT-F1
& BERT-P & BERT-F1
& Acc. & Macro-F1
& Acc.
& Mean \\
\midrule

\multirow{3}{*}{PTMs}
& MAE
& 53.20 & 51.33
& -- & --
& -- & --
& 55.30 & 51.59
& --
& -- \\

& ViT
& 54.80 & 52.88
& -- & --
& -- & --
& 59.30 & 54.12
& --
& -- \\

& CLIP
& \textbf{54.90} & \textbf{53.00}
& -- & --
& -- & --
& \textbf{69.70} & \textbf{65.28}
& --
& -- \\

\midrule

\multirow{2}{*}{\shortstack[l]{Task-\\specific}}
& C4MMD
& 56.40 & 54.22
& -- & --
& -- & --
& -- & --
& --
& -- \\

& SEMD
& \textbf{59.60} & \textbf{57.58}
& -- & --
& -- & --
& \textbf{62.30} & \textbf{57.53}
& --
& -- \\

\midrule

\multirow{5}{*}{\shortstack[l]{Open-weight\\MLLMs}}
& Qwen3-VL-8B
& 57.00 & 52.23
& 67.74 & \textbf{67.89}
& 66.09 & 66.40
& 55.40 & 48.22
& 10.80
& 37.55 \\

& InternVL3.5-14B-HF
& 61.90 & 60.39
& 59.73 & 59.70
& 58.08 & 58.32
& \textbf{69.70} & 60.23
& 16.10
& 36.50 \\

& Kimi-K2.5
& 58.80 & 55.12
& 61.66 & 63.17
& 62.09 & 63.51
& 66.90 & 52.96
& 12.30
& \textbf{48.80} \\

& InternVL3.5-8B-HF
& 62.70 & 60.31
& 64.08 & 64.25
& 63.00 & 63.32
& 66.60 & \textbf{61.01}
& 16.20
& 35.66 \\

& Qwen3-VL-32B
& \textbf{63.50} & \textbf{61.83}
& \textbf{68.03} & 67.54
& \textbf{67.72} & \textbf{66.53}
& 60.30 & 52.49
& \textbf{16.80}
& 37.72 \\

\midrule

\multirow{3}{*}{\shortstack[l]{Proprietary\\MLLMs}}
& Claude-Sonnet-4.6
& 52.90 & 39.03
& 69.36 & 72.09
& 68.65 & \textbf{71.35}
& 75.00 & 57.11
& 12.40
& \textbf{49.77} \\

& GPT-5.5
& \textbf{59.30} & \textbf{54.56}
& 67.13 & 68.74
& 65.37 & 67.02
& \textbf{75.70} & 61.77
& \textbf{18.40}
& 37.76 \\

& GPT-4o
& 58.10 & 50.19
& \textbf{72.64} & \textbf{72.39}
& \textbf{71.37} & 71.23
& 75.50 & \textbf{64.69}
& 13.50
& 40.71 \\

\bottomrule
\end{tabular*}

\caption{Experiment results on \MthreeRBench{}. Rubric is the mean of ITE, MC, and SJ; "-" denotes unsupported outputs; the best result within each model type is \textbf{bold}.}
\label{tab:benchmark_results}

\end{table*}

\section{Performance on \MthreeRBench{}}
\label{sec:benchmark-performance}

\subsection{Evaluation Setting}

\paragraph{Task Definition.}
\label{sec:task-definition}
Given an image--text pair $x=(I,T)$, a system predicts
$y=(m,t,s,e)$: metaphor occurrence $m\in\{\text{Yes},\text{No}\}$, the most salient Target $t$ and Source $s$, and
sentiment
$e\in\{\text{Positive},\text{Neutral},\text{Negative}\}$. A metaphor-occurrence label of Yes requires a directional cross-domain
mapping grounded in the image,
text, or both; otherwise $t=s=\texttt{None}$. These fields
define Metaphor Detection (MD), Target Prediction (TP), Source Prediction
(SP), and Sentiment Classification (SC). The system also generates stage-wise
rationales for Image-Text Evidence (ITE), Mapping Correctness (MC), and Sentiment
Justification (SJ), with MC evaluated only on gold metaphors. {\color{red}

\paragraph{Evaluation Metrics.}
MD and SC use Accuracy and Macro-F1 over all $N=1{,}000$ instances. TP and SP
use mean BERTScore Precision and F1~\cite{zhang2020bertscore} over the fixed
set of $N_M=510$ gold metaphors; a missed metaphor receives zero for both
mapping fields. Gold non-metaphors are excluded from TP/SP
because no gold mapping is defined for them; metaphor false positives are
instead penalized by MD and JLA. To measure end-to-end success, Joint Label Accuracy (JLA)
requires all four fields to be correct on the same instance. Let $c_i^t$ and
$c_i^s$ indicate Target and Source correctness: BERTScore F1 must reach the
calibrated threshold $0.75$ for a gold metaphor, whereas both fields must be
\texttt{None} for a gold non-metaphor. We compute:
\begin{equation}
\mathrm{JLA}=\frac{1}{N}\sum_{i=1}^{N}
\mathbb{I}[\hat m_i=m_i]c_i^t c_i^s
\mathbb{I}[\hat e_i=e_i].
\label{eq:jla}
\end{equation}

Unless noted, performance metrics are reported a 0-100 scale. For explanation quality,
Gemini-3.1-Pro at temperature 0 compares each predicted stage with its
human-verified reference. ITE, MC, or SJ receives zero when its prerequisite
metaphor, mapping, or sentiment prediction is incorrect; otherwise the judge
scores semantic agreement.
Appendix~\ref{app:evaluation} specifies the BERTScore configuration, parsing,
gates, and edge cases; Appendix~\ref{app:judge} describes the reference-based
judge protocol.
}

\paragraph{Evaluated Models.}
We evaluate representative models from four categories: general-purpose
pre-trained models (CLIP~\cite{radford2021clip}, ViT~\cite{dosovitskiy2021image},
and MAE~\cite{he2022masked}), task-specific metaphor models
(C4MMD~\cite{xu2024exploring} and SEMD~\cite{yang2025multimm}), open-weight
MLLMs (Qwen3-VL-8B, Qwen3-VL-32B~\cite{bai2025qwen3vl},
InternVL3.5-8B-HF, InternVL3.5-14B-HF~\cite{wang2025internvl35}, and
Kimi-K2.5~\cite{moonshotai2026kimik25}), and proprietary MLLMs
(GPT-4o~\cite{openai2024gpt4o}, GPT-5.5~\cite{openai2026gpt55}, and
Claude-Sonnet-4.6~\cite{anthropic2026claudesonnet46}).

\begin{figure*}[t]
\centering
\includegraphics[
width=0.98\textwidth,
]{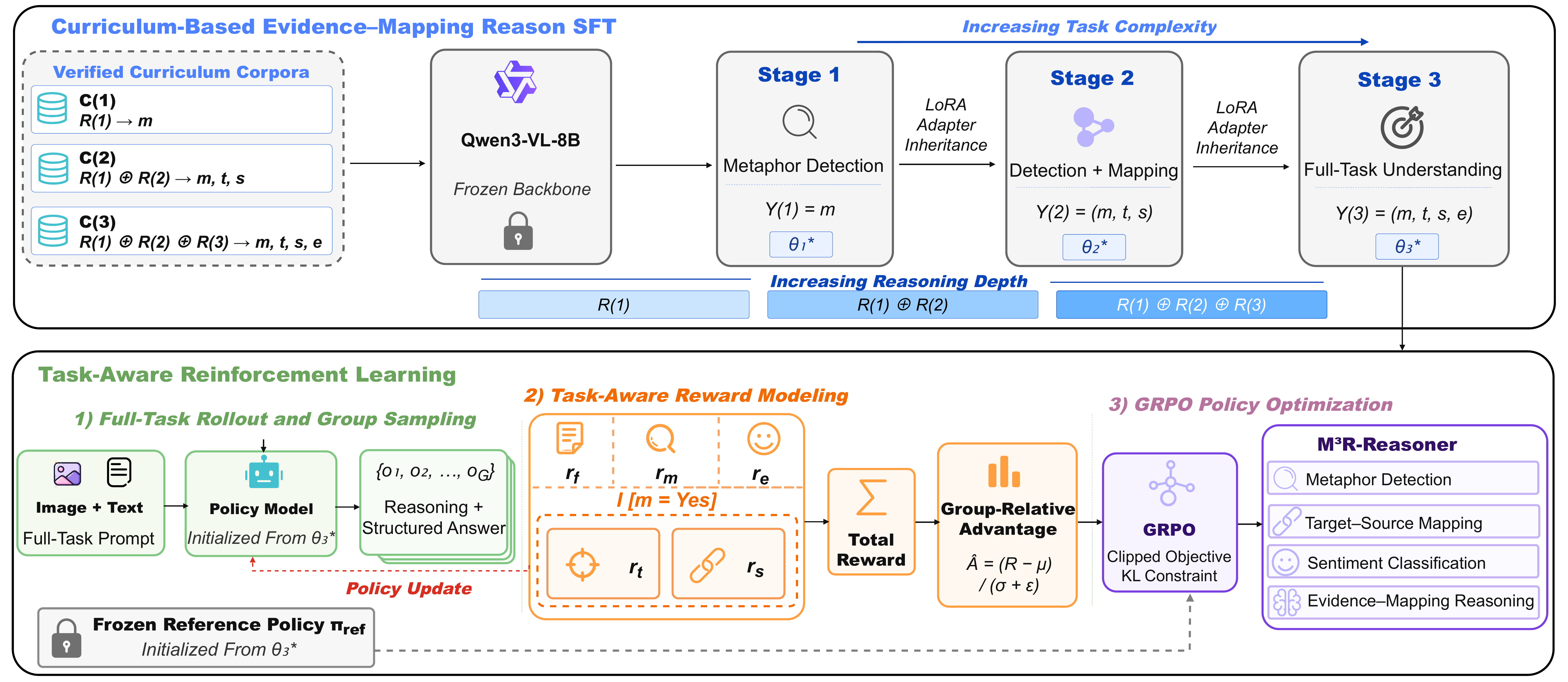}
\caption{Training flow of \MthreeRReasoner{}. (a) Curriculum SFT transfers LoRA adapters across increasingly complex evidence--mapping supervision stages. (b) Task-aware reinforcement learning scores full-task rollouts with structured rewards and applies GRPO against a frozen reference policy.} \label{fig:method}
\end{figure*}

\subsection{Experiment Results}

We explore the following two research questions.

%
\paragraph{RQ1: How well do existing models perform on unified
evidence-grounded multimodal metaphor understanding?}

Table~\ref{tab:benchmark_results} reports performance on
\MthreeRBench{}. We find: (1) Non-MLLM methods remain limited. CLIP reaches
54.90\% Accuracy and 53.00\% Macro-F1 for metaphor detection, while SEMD
improves to 59.60\% Accuracy and 57.58\% Macro-F1 but cannot generate
open-ended Target--Source predictions. (2) MLLMs perform better but unevenly
across tasks. Qwen3-VL-32B obtains the best metaphor-detection Accuracy
(63.50\%), GPT-4o the best Target prediction (72.39\% BERTScore F1), and
GPT-5.5 the highest sentiment Accuracy (75.70\%). Yet GPT-5.5 achieves only
18.40\% JLA, so fewer than one in five instances receive a fully correct
four-field prediction. This component--joint gap highlights the difficulty of
unified evidence-grounded metaphor understanding.

\paragraph{RQ2: What are the primary sources of error in existing MLLMs?}

For error analysis, we randomly sample 100 GPT-4o errors. We select GPT-4o
because it attains the best result on more component-level metrics than any
other full-task baseline. We find that \emph{Textual Shortcut Bias} is the
dominant failure mode (43\%), where the model relies on slogans, exaggerations,
or promotional language while ignoring visual evidence.
\emph{Target--Source Granularity Mismatch} accounts for 26\%, with the model
predicting abstract themes rather than the entities involved in metaphorical
mappings. \emph{Visual Grounding Failure} contributes another 21\%, especially
for metaphors involving entity substitution, morphological fusion, spatial
embedding, or symbolic transformation. The remaining errors comprise
sentiment misclassification (8\%) and hallucinated reasoning (2\%). Overall,
the first three categories account for 90\% of failures, showing that
insufficient evidence grounding and conceptual alignment remain the primary
challenges for current MLLMs.

\section{\MthreeRReasoner{}}
\label{sec:m3reasoner}

{\color{red}
Motivated by the findings above, we propose \MthreeRReasoner{} for multimodal
metaphor understanding. As illustrated in Figure~\ref{fig:method}, the pipeline
first constructs a reasoning supervision corpus and then applies two training
stages: curriculum-based evidence--mapping SFT and task-aware reinforcement
learning.
}

{\color{red}
\subsection{Reasoning Supervision Corpus Construction}
\label{sec:reasoning_corpus}
Five MLLMs independently annotate each candidate using the structured schema
and complete-tuple agreement rule in Section~\ref{sec:m3rbench}. We retain a
candidate only when the same subset of at least three models agrees on all
applicable fields. For each retained candidate, we generate stage-wise
explanations, verify them against the agreed labels, and discard malformed
outputs, unsupported inferences, and label--explanation inconsistencies. This
pipeline yields 8,989 training instances, which we organize into three corpora
of increasing complexity: $\mathcal{C}^{(1)}$ covers MD and evidence-grounded
justification; $\mathcal{C}^{(2)}$ adds TP, SP, and attribute projection; and
$\mathcal{C}^{(3)}$ adds SC and sentiment justification.
Appendix~\ref{app:reasoning-corpus} details the quality-control and output-format
rules.
}
{\color{red}
\subsection{Curriculum-Based Evidence--Mapping Supervised Fine-tuning}
\label{sec:curriculum_sft}
}

We adopt Qwen3-VL-8B~\cite{bai2025qwen3vl} as the base model
and perform supervised fine-tuning sequentially on
$\mathcal{C}^{(1)} \rightarrow \mathcal{C}^{(2)}
\rightarrow \mathcal{C}^{(3)}$.
This curriculum progressively expands the prediction space and
evidence-grounded reasoning requirements, allowing each stage to build
on the capabilities acquired previously.

Across all stages, the backbone parameters are frozen and only the
LoRA adapters~\cite{hu2022lora} are optimized.
For $k=1,2,3$, we update the adapters sequentially as follows:
\begin{equation}
\theta_k^{\star}=\operatorname{SFT}(\theta_{k-1}^{\star};\mathcal{C}^{(k)}),
\quad \theta_0^{\star}=\theta_0.
\label{eq:curriculum_sft}
\end{equation}
where $\mathcal{C}^{(k)}$ denotes the stage-$k$ corpus and
$\theta_k^{\star}$ denotes the resulting model state.
The intermediate states $\theta_1^{\star}$ and
$\theta_2^{\star}$ transfer the progressively acquired capabilities
to subsequent stages, while $\theta_3^{\star}$ initializes the
reinforcement-learning stage.

\subsection{Task-Aware Reinforcement Learning}
\label{sec:task_aware_rl}

After curriculum-based SFT, we initialize the policy $\pi_\theta$ from
$\theta_3^{\star}$ and jointly optimize MD, TP, SP, and SC. For input $x_i$,
the policy samples $G$ responses $\{o_{i,g}\}_{g=1}^{G}$ and scores each as follows:
\begin{equation}
\begin{aligned}
R_{i,g}={}&
\lambda_f r_f+\lambda_m r_m+\lambda_e r_e \\
&+\mathbb{I}[m_i=\texttt{Yes}]
\left(\lambda_t r_t+\lambda_s r_s\right),
\end{aligned}
\label{eq:full_task_reward}
\end{equation}
where $r_f$ checks output-format validity, while $r_m$ and $r_e$ score exact
matches of the normalized metaphor and sentiment labels. For metaphorical samples,
$r_t$ and $r_s$ are the BERTScore F1 values between the predicted and gold
Target and Source~\cite{zhang2020bertscore}. The indicator omits mapping rewards for
non-metaphorical samples; any invalid or missing field receives zero.

We optimize $\pi_\theta$ with Group Relative Policy Optimization
(GRPO)~\cite{shao2024deepseekmath}, which derives relative advantages from the
$G$ rewards and applies clipped policy updates. Both $\pi_\theta$ and the
frozen reference policy $\pi_{\mathrm{ref}}$ are initialized from
$\theta_3^{\star}$; $\pi_{\mathrm{ref}}$ supplies KL regularization to preserve
the evidence-grounded behavior acquired during SFT.

\section{\MthreeRReasoner{} Evaluation}
\label{sec:m3reasoner_evaluation}

\paragraph{Implementation Details.}
We train \MthreeRReasoner{} from Qwen3-VL-8B with a frozen backbone and
rank-8 LoRA adapters, then evaluate it on \MthreeRBench{}. Curriculum SFT runs on
$\mathcal{C}^{(1)}$, $\mathcal{C}^{(2)}$, and $\mathcal{C}^{(3)}$; GRPO then
starts from $\theta_3^{\star}$ and samples 4 responses per prompt.
Appendix~\ref{app:training-controls} provides the detailed implementation settings.

\begin{table}[t]
\centering

\small
\setlength{\tabcolsep}{3.2pt}

\begin{tabular*}{\columnwidth}{
@{\extracolsep{\fill}}
lccccc
@{}
}
\toprule
Model & MD & TP & SP & SC & JLA \\
\midrule

\multicolumn{6}{@{}l}{\textit{Proprietary MLLMs}} \\
\midrule

GPT-4o
& 50.19 & 72.39 & 71.23 & 64.69 & 13.50 \\

GPT-5.5
& 54.56 & 68.74 & 67.02 & 61.77 & 18.40 \\

Claude-Sonnet-4.6
& 39.03 & 72.09 & 71.35 & 57.11 & 12.40 \\

\midrule

\multicolumn{6}{@{}l}{\textit{Open-Weight MLLMs}} \\
\midrule

Qwen3-VL-8B
& 52.23 & 67.89 & 66.40 & 48.22 & 10.80 \\

InternVL3.5-8B-HF
& 60.31 & 64.25 & 63.32 & 61.01 & 16.20 \\

InternVL3.5-14B-HF
& 60.39 & 59.70 & 58.32 & 60.23 & 16.10 \\

Qwen3-VL-32B
& 61.83 & 67.54 & 66.53 & 52.49 & 16.80 \\

Kimi-K2.5
& 55.12 & 63.17 & 63.51 & 52.96 & 12.30 \\

\midrule

{\MthreeRReasoner{} (Ours)}
& \textbf{72.53}
& \textbf{75.74}
& \textbf{73.31}
& \textbf{71.10}
& \textbf{37.90} \\

\bottomrule
\end{tabular*}

\caption{Main results on \MthreeRBench{}. MD and SC report Macro-F1; TP
and SP report BERTScore F1.}
\label{tab:m3reasoner_main}

\end{table}

\paragraph{Experiment Results.}

Because no existing task-specific baseline supports the complete open-ended
output required by our unified setting, we compare \MthreeRReasoner{} with
representative proprietary and open-weight MLLMs. As shown in
Table~\ref{tab:m3reasoner_main}, \MthreeRReasoner{} achieves the best result on
every component metric, improving MD, TP, SP, and SC over Qwen3-VL-8B by
20.30, 7.85, 6.91, and 22.88 points, respectively. Its JLA reaches 37.90\%,
exceeding the base model by 27.10 points and the strongest proprietary
baseline, GPT-5.5, by 19.50 points. The gains in both component metrics and
JLA show that the proposed training pipeline improves not only individual
subtasks but also their joint execution.

\begin{table}[t]
\centering

\small
\setlength{\tabcolsep}{1.5pt}

\begin{tabular*}{\columnwidth}{
@{\extracolsep{\fill}}
lccccc
@{}
}
\toprule
Setting & MD & TP & SP & SC & JLA \\
\midrule

Qwen3-VL-8B
& 52.23
& 67.89
& 66.40
& 48.22
& 10.80 \\

Label-SFT + Curr.
& 65.37
& 70.88
& 68.44
& 70.66
& 36.70 \\

CoT-SFT w/o Curr.
& 66.82
& 70.70
& 68.57
& 69.85
& 36.80 \\

CoT-SFT + Curr.
& 71.90
& 74.58
& 72.58
& 70.72
& 37.50\\

\midrule

\MthreeRReasoner{} w/o {\normalsize $r_{ts}$}
& 66.27
& 72.90
& 70.75
& 69.91
& 36.30 \\

\MthreeRReasoner{} w/o {\normalsize $r_e$}
& 67.44
& 69.86
& 67.58
& 70.26
& 37.10 \\

\MthreeRReasoner{}
& \textbf{72.53}
& \textbf{75.74}
& \textbf{73.31}
& \textbf{71.10}
& \textbf{37.90} \\

\bottomrule
\end{tabular*}

\caption{Ablation results on \MthreeRBench{}.
MD and SC report Macro-F1, TP and SP report BERTScore F1.} \label{tab:m3reasoner_ablation}

\end{table}

\paragraph{Ablation Study.}

Table~\ref{tab:m3reasoner_ablation} reports ablations of
evidence-grounded CoT, curriculum learning, and task-aware reinforcement
learning; $r_{ts}$ abbreviates the paired Target/Source rewards $r_t+r_s$. Compared with
label-only curriculum SFT, CoT yields selective rather than uniform gains:
MD, TP, and SP improve by 6.53, 3.70, and 4.14 points, whereas SC changes by
only 0.06 points and JLA increases modestly from 36.70\% to 37.50\%. Its main
additional value is an explicit evidence--mapping--sentiment trace that makes
intermediate errors inspectable. Relative to non-curriculum CoT-SFT, the
curriculum further improves MD, TP, and SP by 5.08, 3.88, and 4.01 points.
Task-aware reinforcement learning achieves the strongest overall
results and raises JLA to 37.90\%; removing either $r_{ts}$ or $r_e$
consistently degrades performance.

\paragraph{Prompting Baselines.}
We test whether stronger task-specific prompts improve general-purpose MLLMs or account for \MthreeRReasoner{}'s gains. We compare Direct, Definitions, and Evidence CoT prompts for GPT-5.5, the strongest proprietary baseline, and Qwen3-VL-8B, the \MthreeRReasoner{} backbone. Within each model, prompts share inputs, decoding, parsing, and evaluation; \MthreeRReasoner{} uses Direct prompting at inference. Appendix~\ref{app:prompting} provides the full prompts. \begin{table}[t]
\centering
\small
\setlength{\tabcolsep}{1.2pt}
\begin{tabular*}{\columnwidth}{@{\extracolsep{\fill}}llccccc@{}}
\toprule
Model & Prompt & MD & TP & SP & SC & JLA \\
\midrule
\multirow{3}{*}{GPT-5.5}
& Direct & 54.56 & 68.74 & 67.02 & 61.77 & 18.40 \\
& Definitions & 56.14 & 68.53 & 65.06 & 59.80 & 23.00 \\
& Evidence CoT & 56.62 & 68.07 & 65.17 & 60.48 & 23.00 \\
\addlinespace[0.5pt]
\multirow{3}{*}{Qwen3-VL-8B}
& Direct & 52.23 & 67.89 & 66.40 & 48.22 & 10.80 \\
& Definitions & 50.36 & 68.77 & 67.50 & 59.80 & 10.60 \\
& Evidence CoT & 49.38 & 71.32 & 69.41 & 59.06 & 10.00 \\
\midrule
\MthreeRReasoner{} & Direct & \textbf{72.53} & \textbf{75.74} &
\textbf{73.31} & \textbf{71.10} & \textbf{37.90} \\
\bottomrule
\end{tabular*}
\caption{Controlled prompting results on \MthreeRBench{}. MD and SC report Macro-F1, TP and SP report
BERTScore F1, and JLA reports joint-label accuracy.} \label{tab:prompting_controls}
\end{table}

Table~\ref{tab:prompting_controls} shows limited and inconsistent gains from stronger prompts. For GPT-5.5, they improve MD by at most 2.06 points and JLA by 4.60 points, but not TP, SP, or SC. For Qwen3-VL-8B, they improve TP, SP, and SC but reduce MD and do not improve JLA. By contrast, \MthreeRReasoner{} reaches 37.90 JLA, exceeding the GPT-5.5 and Qwen3-VL-8B Evidence CoT variants by 14.90 and 27.90 points while leading every component. Thus, prompting benefits selected subtasks but does not account for the overall gains of \MthreeRReasoner{}.

\begin{table}[t]
\centering
\small
\setlength{\tabcolsep}{1.5pt}
\begin{tabular*}{\columnwidth}{@{\extracolsep{\fill}}lcccc@{}}
\toprule
Model & ITE & MC & SJ & Mean \\
\midrule
Qwen3-VL-8B & 38.63 & 31.61 & 42.40 & 37.55 \\
InternVL3.5-14B-HF & 36.60 & 25.42 & 47.48 & 36.50 \\
GPT-5.5 & 29.66 & 43.95 & 39.68 & 37.76 \\
GPT-4o & 38.53 & 36.58 & 47.01 & 40.71 \\
Claude-Sonnet-4.6 & 43.56 & 43.90 & 61.84 & 49.77 \\
\addlinespace[0.5pt]
CoT-SFT + Curr. & 58.01 & 42.30 & 67.52 & 55.94 \\
\MthreeRReasoner{} & \textbf{58.11} & \textbf{44.41} &
\textbf{67.79} & \textbf{56.77} \\
\bottomrule
\end{tabular*}
\caption{Evidence-grounded rubric results on \MthreeRBench{}. Mean averages ITE, MC, and SJ.}
\label{tab:rubric_results}
\end{table}

\paragraph{Rubric-Based Explanation Evaluation.}
Two annotators independently score applicable ITE, MC, and SJ outputs from Claude-Sonnet-4.6, CoT-SFT + Curr., and \MthreeRReasoner{} on a stratified 200-instance subset; their per-item means are then compared with Gemini scores. Annotators are blinded to model identities and Gemini scores. The scores are strongly correlated, differ by at most 4\% of the scale, and yield identical system rankings across dimensions. We therefore use Gemini for the label-gated full-benchmark results in Table~\ref{tab:rubric_results}; Appendix~\ref{app:metric-validation} gives the protocol and agreement statistics. \MthreeRReasoner{} achieves 58.11 ITE, 44.41 MC, and 67.79 SJ,
yielding the highest mean of 56.77. Compared with Claude-Sonnet-4.6, the
strongest proprietary baseline in this evaluation, it improves ITE, MC, SJ,
and Mean by 14.55, 0.51, 5.95, and 7.00 points, respectively. It also
consistently outperforms CoT-SFT + Curr., showing that task-aware
reinforcement learning further improves image-text evidence references,
Target--Source mapping, and sentiment justification.

\begin{figure}[t]
\centering
\includegraphics[width=\columnwidth]{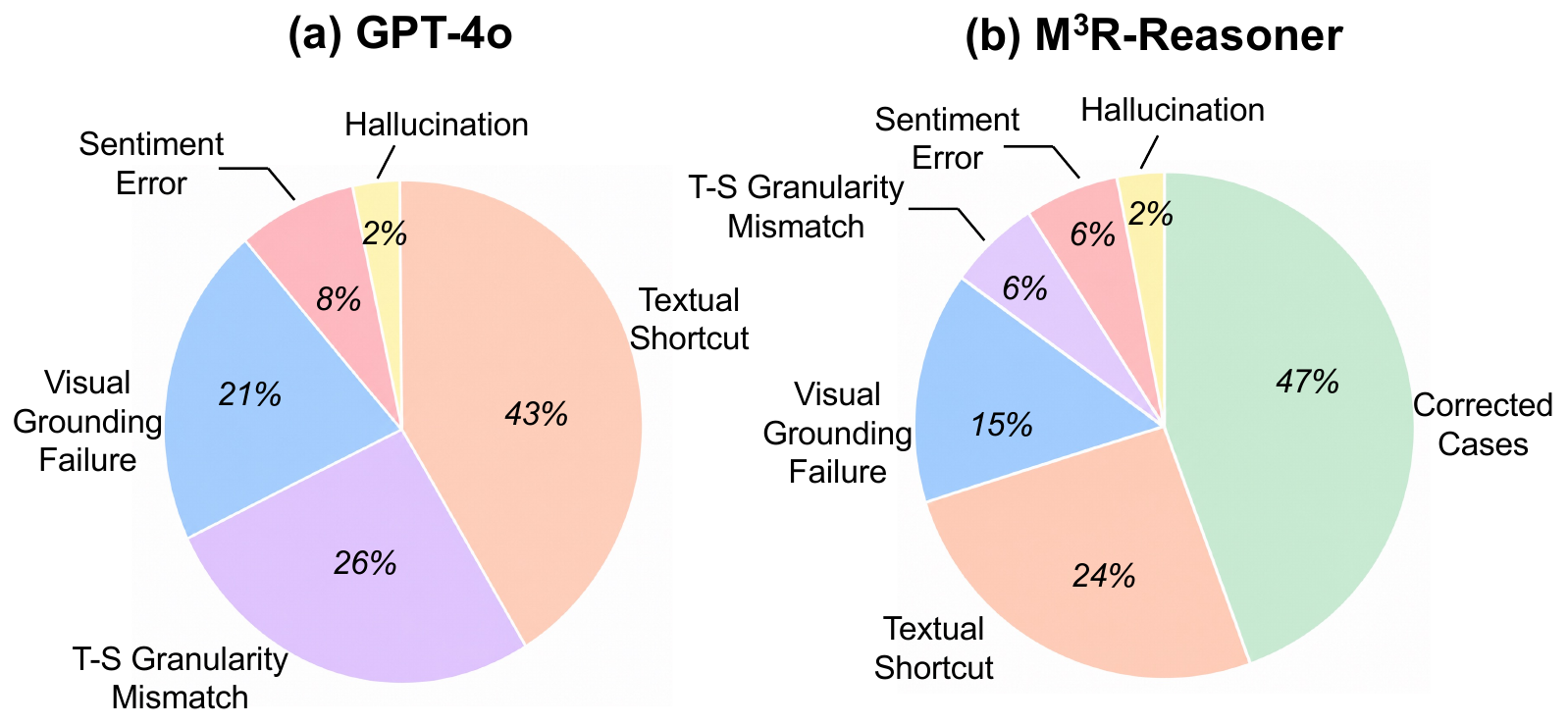}
\caption{Error distributions of GPT-4o and \MthreeRReasoner{} on the same
100 GPT-4o failure cases.}
\label{fig:error_analysis}
\end{figure}

\paragraph{Error Analysis.}
Figure~\ref{fig:error_analysis} compares \MthreeRReasoner{} with GPT-4o on the
100 erroneous cases from RQ2. \MthreeRReasoner{} corrects 47\% of these
failures, with the largest improvements in Target--Source granularity mismatch
(reduced from 26 to 6 cases) and textual shortcut bias (from 43 to 24 cases).
Visual grounding failures decrease from 21 to 15 cases, while sentiment errors
drop from 8 to 6. Hallucinated reasoning remains unchanged, suggesting a
broader limitation of current VLMs. Overall, evidence-grounded training
improves conceptual mapping and reduces textual shortcuts, while visual
grounding remains a significant challenge. Appendix~\ref{app:qualitative-analysis} presents a concrete case study.

\section{Conclusion}

We introduced \MthreeRBench{}, a unified evidence-grounded benchmark that
jointly evaluates metaphor identification, Target--Source mapping, and
sentiment inference with stage-wise reference explanations. Evaluations on
\MthreeRBench{} reveal that existing models frequently overlook visual
evidence, rely on superficial textual cues, and produce inaccurate conceptual
mappings. To address these limitations, we proposed \MthreeRReasoner{}, which
combines curriculum-based reasoning supervision with task-aware
reinforcement learning. Experiments show that \MthreeRReasoner{} achieves the
best overall performance across all four task metrics and produces explanations
that are more consistent with human-verified image-text evidence and metaphorical
mappings. We hope this benchmark and method provide a foundation for advancing
interpretable multimodal metaphor understanding.

\section*{Limitations}

\MthreeRBench{} contains 1,000 carefully annotated image--text pairs, but the
benchmark remains limited in scale. Future work will expand it with
additional human-verified instances. Despite the gains of \MthreeRReasoner{},
fine-grained identification of metaphor-relevant image-text evidence and end-to-end
performance measured by Joint Label Accuracy (JLA) leave substantial room for
improvement. Finally, modeling and evaluating multiple Target--Source mappings
within a single multimodal metaphor instance is an important direction for
future work.

\section*{Ethical Considerations}

\MthreeRBench{} comprises image--text pairs drawn from six previously released resources and re-annotated for research on multimodal metaphor understanding. Because these resources include social-media posts and advertisements, they may contain culturally specific associations, stereotypes, sensitive imagery, or depictions of identifiable individuals. Human verification improves annotation quality but cannot eliminate subjective or cultural bias in metaphor and sentiment interpretation. The three primary annotators received standardized training and compensation, as detailed in Appendix~\ref{app:protocol}. Any access to or redistribution of image content should comply with the licenses and terms of the original resources.

\bibliography{m3r_bench}

@book{lakoff1980metaphors,
  author    = {George Lakoff and Mark Johnson},
  title     = {Metaphors We Live By},
  year      = {1980},
  publisher = {University of Chicago Press},
  address   = {Chicago}
}

@book{forceville2009multimodal,
  editor    = {Charles J. Forceville and Eduardo Urios-Aparisi},
  title     = {Multimodal Metaphor},
  year      = {2009},
  publisher = {Mouton de Gruyter},
  address   = {Berlin and New York},
  series    = {Applications of Cognitive Linguistics},
  volume    = {11},
  doi       = {10.1515/9783110215366}
}

@inproceedings{zhang2021multimet,
  author    = {Dongyu Zhang and Minghao Zhang and Heting Zhang and Liang Yang and Hongfei Lin},
  title     = {{MultiMET}: A Multimodal Dataset for Metaphor Understanding},
  booktitle = {Proceedings of the 59th Annual Meeting of the Association for Computational Linguistics and the 11th International Joint Conference on Natural Language Processing (Volume 1: Long Papers)},
  year      = {2021},
  pages     = {3214--3225},
  publisher = {Association for Computational Linguistics},
  address   = {Online},
  doi       = {10.18653/v1/2021.acl-long.249},
  url       = {https://aclanthology.org/2021.acl-long.249/}
}

@inproceedings{xu2022metmeme,
  author    = {Bo Xu and Tingting Li and Junzhe Zheng and Mehdi Naseriparsa and Zhehuan Zhao and Hongfei Lin and Feng Xia},
  title     = {{MET-Meme}: A Multimodal Meme Dataset Rich in Metaphors},
  booktitle = {Proceedings of the 45th International ACM SIGIR Conference on Research and Development in Information Retrieval},
  year      = {2022},
  pages     = {2887--2899},
  publisher = {Association for Computing Machinery},
  doi       = {10.1145/3477495.3532019},
  url       = {https://doi.org/10.1145/3477495.3532019}
}

@inproceedings{zhang2023multicmet,
  author    = {Dongyu Zhang and Jingwei Yu and Senyuan Jin and Liang Yang and Hongfei Lin},
  title     = {{MultiCMET}: A Novel Chinese Benchmark for Understanding Multimodal Metaphor},
  booktitle = {Findings of the Association for Computational Linguistics: EMNLP 2023},
  year      = {2023},
  pages     = {6141--6154},
  publisher = {Association for Computational Linguistics},
  address   = {Singapore},
  doi       = {10.18653/v1/2023.findings-emnlp.409},
  url       = {https://aclanthology.org/2023.findings-emnlp.409/}
}

@inproceedings{yang2025multimm,
  author    = {Senqi Yang and Dongyu Zhang and Jing Ren and Ziqi Xu and Xiuzhen Zhang and Yiliao Song and Hongfei Lin and Feng Xia},
  title     = {Cultural Bias Matters: A Cross-Cultural Benchmark Dataset and Sentiment-Enriched Model for Understanding Multimodal Metaphors},
  booktitle = {Proceedings of the 63rd Annual Meeting of the Association for Computational Linguistics (Volume 1: Long Papers)},
  year      = {2025},
  pages     = {26301--26317},
  publisher = {Association for Computational Linguistics},
  address   = {Vienna, Austria},
  doi       = {10.18653/v1/2025.acl-long.1275},
  url       = {https://aclanthology.org/2025.acl-long.1275/}
}

@inproceedings{lu2025emometa,
  author    = {Xingyuan Lu and Yuxi Liu and Dongyu Zhang and Zhiyao Wu and Jing Ren and Feng Xia},
  title     = {{EmoMeta}: A Multimodal Dataset for Fine-Grained Emotion Classification in Chinese Metaphors},
  booktitle = {Companion Proceedings of the ACM on Web Conference 2025},
  year      = {2025},
  pages     = {3080--3083},
  publisher = {Association for Computing Machinery},
  doi       = {10.1145/3701716.3735080},
  url       = {https://doi.org/10.1145/3701716.3735080}
}

@article{zhang2025cm3d,
  author  = {Dongyu Zhang and Shengcheng Yin and Jingwei Yu and Zhiyao Wu and Zhen Li and Chengpei Xu and Xiaoxia Wang and Feng Xia},
  title   = {Towards Multimodal Metaphor Understanding: A Chinese Dataset and Model for Metaphor Mapping Identification},
  journal = {ACM Transactions on Asian and Low-Resource Language Information Processing},
  year    = {2025},
  volume  = {24},
  number  = {12},
  pages   = {1--25},
  doi     = {10.1145/3773989},
  url     = {https://doi.org/10.1145/3773989}
}

@inproceedings{akula2023metaclue,
  author    = {Arjun R. Akula and others},
  title     = {{MetaCLUE}: Towards Comprehensive Visual Metaphors Research},
  booktitle = {Proceedings of the IEEE/CVF Conference on Computer Vision and Pattern Recognition},
  year      = {2023},
  pages     = {23201--23211},
  url       = {https://openaccess.thecvf.com/content/CVPR2023/html/Akula_MetaCLUE_Towards_Comprehensive_Visual_Metaphors_Research_CVPR_2023_paper.html}
}

@inproceedings{kundu2025imagemet,
  author    = {Manishit Kundu and Sumit Shekhar and Pushpak Bhattacharyya},
  title     = {Looking Beyond the Pixels: Evaluating Visual Metaphor Understanding in {VLM}s},
  booktitle = {Findings of the Association for Computational Linguistics: EMNLP 2025},
  year      = {2025},
  pages     = {23137--23158},
  publisher = {Association for Computational Linguistics},
  address   = {Suzhou, China},
  doi       = {10.18653/v1/2025.findings-emnlp.1257},
  url       = {https://aclanthology.org/2025.findings-emnlp.1257/}
}

@inproceedings{zheng2026m3ucd,
  author    = {Tianlong Zheng and Yating Yang and Rui Dong and Bo Ma and Lei Wang and Xi Zhou and Siru Miao and Turghun Osman},
  title     = {{M3UCD}: A Multi-Task Multimodal Metaphor Understanding Challenge Dataset for {LLM}s},
  booktitle = {Proceedings of the AAAI Conference on Artificial Intelligence},
  year      = {2026},
  volume    = {40},
  number    = {41},
  pages     = {35030--35040},
  doi       = {10.1609/aaai.v40i41.40808},
  url       = {https://doi.org/10.1609/aaai.v40i41.40808}
}

@article{zhang2025coc,
  author  = {Dongyu Zhang and Xingyuan Lu and Mulin Zhuang and Senqi Yang and Hongjun Chen},
  title   = {Multimodal Metaphor Recognition Based on Chain-of-Cognition Prompting},
  journal = {Cognitive Systems Research},
  year    = {2025},
  volume  = {91},
  pages   = {101356},
  doi     = {10.1016/j.cogsys.2025.101356},
  url     = {https://doi.org/10.1016/j.cogsys.2025.101356}
}

@inproceedings{tian2025imara,
  author    = {Yuan Tian and Minzheng Wang and Nan Xu and Wenji Mao},
  title     = {{ImaRA}: An Imaginative Frame Augmented Method for Low-Resource Multimodal Metaphor Detection and Explanation},
  booktitle = {Findings of the Association for Computational Linguistics: NAACL 2025},
  year      = {2025},
  pages     = {3953--3967},
  publisher = {Association for Computational Linguistics},
  address   = {Albuquerque, New Mexico},
  doi       = {10.18653/v1/2025.findings-naacl.220},
  url       = {https://aclanthology.org/2025.findings-naacl.220/}
}

@inproceedings{birke2006trofi,
  author    = {Julia Birke and Anoop Sarkar},
  title     = {A Clustering Approach for Nearly Unsupervised Recognition of Nonliteral Language},
  booktitle = {11th Conference of the European Chapter of the Association for Computational Linguistics},
  year      = {2006},
  pages     = {329--336},
  publisher = {Association for Computational Linguistics},
  address   = {Trento, Italy},
  url       = {https://aclanthology.org/E06-1042/}
}

@book{steen2010mipvu,
  author    = {Gerard J. Steen and Aletta G. Dorst and J. Berenike Herrmann and Anna A. Kaal and Tina Krennmayr and Tryntje Pasma},
  title     = {A Method for Linguistic Metaphor Identification: From {MIP} to {MIPVU}},
  year      = {2010},
  publisher = {John Benjamins},
  address   = {Amsterdam},
  series    = {Converging Evidence in Language and Communication Research},
  volume    = {14},
  doi       = {10.1075/celcr.14}
}

@inproceedings{shao2024cmdag,
  author    = {Yujie Shao and Xinrong Yao and Xingwei Qu and Chenghua Lin and Shi Wang and Wenhao Huang and Ge Zhang and Jie Fu},
  title     = {{CMDAG}: A Chinese Metaphor Dataset with Annotated Grounds as {CoT} for Boosting Metaphor Generation},
  booktitle = {Proceedings of the 2024 Joint International Conference on Computational Linguistics, Language Resources and Evaluation (LREC-COLING 2024)},
  year      = {2024},
  pages     = {3357--3366},
  publisher = {ELRA and ICCL},
  address   = {Torino, Italia},
  url       = {https://aclanthology.org/2024.lrec-main.298/}
}

@inproceedings{liu2022figmemes,
  author    = {Chen Liu and Gregor Geigle and Robin Krebs and Iryna Gurevych},
  title     = {{FigMemes}: A Dataset for Figurative Language Identification in Politically-Opinionated Memes},
  booktitle = {Proceedings of the 2022 Conference on Empirical Methods in Natural Language Processing},
  year      = {2022},
  pages     = {7069--7086},
  publisher = {Association for Computational Linguistics},
  address   = {Abu Dhabi, United Arab Emirates},
  doi       = {10.18653/v1/2022.emnlp-main.476},
  url       = {https://aclanthology.org/2022.emnlp-main.476/}
}

@inproceedings{zhang2025ciibench,
  author    = {Chenhao Zhang and others},
  title     = {Can {MLLM}s Understand the Deep Implication Behind Chinese Images?},
  booktitle = {Proceedings of the 63rd Annual Meeting of the Association for Computational Linguistics (Volume 1: Long Papers)},
  year      = {2025},
  pages     = {14369--14402},
  publisher = {Association for Computational Linguistics},
  address   = {Vienna, Austria},
  doi       = {10.18653/v1/2025.acl-long.700},
  url       = {https://aclanthology.org/2025.acl-long.700/}
}

@article{wilks1975preferential,
  author  = {Yorick Wilks},
  title   = {A Preferential, Pattern-Seeking, Semantics for Natural Language Inference},
  journal = {Artificial Intelligence},
  year    = {1975},
  volume  = {6},
  number  = {1},
  pages   = {53--74},
  doi     = {10.1016/0004-3702(75)90016-8}
}

@inproceedings{zhang2020bertscore,
  author    = {Tianyi Zhang and Varsha Kishore and Felix Wu and Kilian Q. Weinberger and Yoav Artzi},
  title     = {{BERTScore}: Evaluating Text Generation with {BERT}},
  booktitle = {International Conference on Learning Representations},
  year      = {2020},
  url       = {https://openreview.net/forum?id=SkeHuCVFDr}
}

@article{ye2021interpreting,
  author    = {Keren Ye and Narges Honarvar Nazari and James Hahn and Zaeem Hussain and Mingda Zhang and Adriana Kovashka},
  title     = {Interpreting the Rhetoric of Visual Advertisements},
  journal   = {IEEE Transactions on Pattern Analysis and Machine Intelligence},
  year      = {2021},
  volume    = {43},
  number    = {4},
  pages     = {1308--1323},
  publisher = {IEEE},
  doi       = {10.1109/TPAMI.2019.2947440}
}

@inproceedings{radford2021clip,
  author    = {Alec Radford and others},
  title     = {Learning Transferable Visual Models from Natural Language Supervision},
  booktitle = {Proceedings of the 38th International Conference on Machine Learning},
  year      = {2021},
  volume    = {139},
  series    = {Proceedings of Machine Learning Research},
  pages     = {8748--8763},
  publisher = {PMLR},
  url       = {https://proceedings.mlr.press/v139/radford21a.html}
}

@inproceedings{dosovitskiy2021image,
  author    = {Alexey Dosovitskiy and others},
  title     = {An Image Is Worth 16x16 Words: Transformers for Image Recognition at Scale},
  booktitle = {International Conference on Learning Representations},
  year      = {2021},
  url       = {https://openreview.net/forum?id=YicbFdNTTy}
}

@inproceedings{he2022masked,
  author    = {Kaiming He and Xinlei Chen and Saining Xie and Yanghao Li and Piotr Doll{\'a}r and Ross Girshick},
  title     = {Masked Autoencoders Are Scalable Vision Learners},
  booktitle = {Proceedings of the IEEE/CVF Conference on Computer Vision and Pattern Recognition},
  year      = {2022},
  pages     = {16000--16009},
  url       = {https://openaccess.thecvf.com/content/CVPR2022/html/He_Masked_Autoencoders_Are_Scalable_Vision_Learners_CVPR_2022_paper.html}
}

@inproceedings{xu2024exploring,
  author    = {Yanzhi Xu and Yueying Hua and Shichen Li and Zhongqing Wang},
  title     = {Exploring Chain-of-Thought for Multi-modal Metaphor Detection},
  booktitle = {Proceedings of the 62nd Annual Meeting of the Association for Computational Linguistics (Volume 1: Long Papers)},
  editor    = {Lun-Wei Ku and Andre Martins and Vivek Srikumar},
  year      = {2024},
  month     = aug,
  pages     = {91--101},
  publisher = {Association for Computational Linguistics},
  address   = {Bangkok, Thailand},
  doi       = {10.18653/v1/2024.acl-long.6},
  url       = {https://aclanthology.org/2024.acl-long.6/}
}

@misc{bai2025qwen3vl,
  author       = {Shuai Bai and others},
  title        = {{Qwen3-VL} Technical Report},
  year         = {2025},
  howpublished = {arXiv preprint arXiv:2511.21631},
  url          = {https://arxiv.org/abs/2511.21631}
}

@misc{wang2025internvl35,
  author       = {Weiyun Wang and others},
  title        = {{InternVL3.5}: Advancing Open-Source Multimodal Models in Versatility, Reasoning, and Efficiency},
  year         = {2025},
  howpublished = {arXiv preprint arXiv:2508.18265},
  url          = {https://arxiv.org/abs/2508.18265}
}

@misc{moonshotai2026kimik25,
  author       = {{Kimi Team} and others},
  title        = {{Kimi K2.5}: Visual Agentic Intelligence},
  year         = {2026},
  howpublished = {arXiv preprint arXiv:2602.02276},
  url          = {https://arxiv.org/abs/2602.02276}
}

@misc{openai2024gpt4o,
  author       = {{OpenAI}},
  title        = {{GPT-4o} System Card},
  year         = {2024},
  howpublished = {arXiv preprint arXiv:2410.21276},
  url          = {https://arxiv.org/abs/2410.21276}
}

@misc{openai2026gpt55,
  author       = {{OpenAI}},
  title        = {{GPT-5.5} System Card},
  year         = {2026},
  month        = apr,
  howpublished = {OpenAI System Card},
  note         = {Published April 23, 2026},
  url          = {https://openai.com/index/gpt-5-5-system-card/}
}

@misc{anthropic2026claudesonnet46,
  author       = {{Anthropic}},
  title        = {{Claude Sonnet 4.6} System Card},
  year         = {2026},
  month        = feb,
  howpublished = {Anthropic System Card},
  note         = {Published February 17, 2026},
  url          = {https://www.anthropic.com/claude-sonnet-4-6-system-card}
}

@misc{xai2025grok41,
  author       = {{xAI}},
  title        = {{Grok 4.1} Model Card},
  year         = {2025},
  month        = nov,
  howpublished = {Model Card},
  note         = {Published November 17, 2025},
  url          = {https://data.x.ai/2025-11-17-grok-4-1-model-card.pdf}
}

@misc{anthropic2026claudeopus47,
  author       = {{Anthropic}},
  title        = {{Claude Opus 4.7} System Card},
  year         = {2026},
  month        = apr,
  howpublished = {Anthropic System Card},
  url          = {https://www.anthropic.com/claude-opus-4-7-system-card}
}

@misc{googledeepmind2026gemini31pro,
  author       = {{Google DeepMind}},
  title        = {{Gemini 3.1 Pro} Model Card},
  year         = {2026},
  month        = feb,
  howpublished = {Model Card},
  note         = {Published February 19, 2026},
  url          = {https://deepmind.google/models/model-cards/gemini-3-1-pro/}
}

@inproceedings{hu2022lora,
  author    = {Edward J. Hu and Yelong Shen and Phillip Wallis and Zeyuan Allen-Zhu and Yuanzhi Li and Shean Wang and Lu Wang and Weizhu Chen},
  title     = {Lo{RA}: Low-Rank Adaptation of Large Language Models},
  booktitle = {International Conference on Learning Representations},
  year      = {2022},
  url       = {https://openreview.net/forum?id=nZeVKeeFYf9}
}

@misc{shao2024deepseekmath,
  author       = {Zhihong Shao and Peiyi Wang and Qihao Zhu and Runxin Xu
                  and Junxiao Song and Xiao Bi and Haowei Zhang
                  and Mingchuan Zhang and Y. K. Li and Y. Wu and Daya Guo},
  title        = {{DeepSeekMath}: Pushing the Limits of Mathematical Reasoning in Open Language Models},
  year         = {2024},
  howpublished = {arXiv preprint arXiv:2402.03300},
  url          = {https://arxiv.org/abs/2402.03300}
}

@article{yang2026multimetu,
  author    = {Yang, Senqi and Zhang, Dongyu and Du, Zihang and Li, Haojia and Wang, Guiru and Wu, Qi and Liao, Wanqiu and Hao, Zeqi and Guo, Wei and Pan, Mingshuo and Lin, Hongfei},
  title     = {Multimodal Metaphor Understanding and Generation in {MLLM}s: A Dataset and Reasoning Framework},
  journal   = {Pattern Recognition},
  year      = {2026},
  month     = oct,
  volume    = {178},
  pages     = {113369},
  issn      = {0031-3203},
  doi       = {10.1016/j.patcog.2026.113369},
  url       = {https://doi.org/10.1016/j.patcog.2026.113369}
}

@inproceedings{saakyan2025vflute,
  author    = {Saakyan, Arkadiy and Kulkarni, Shreyas and Chakrabarty, Tuhin and Muresan, Smaranda},
  title     = {Understanding Figurative Meaning through Explainable Visual Entailment},
  editor    = {Chiruzzo, Luis and Ritter, Alan and Wang, Lu},
  booktitle = {Proceedings of the 2025 Conference of the Nations of the Americas Chapter of the Association for Computational Linguistics: Human Language Technologies (Volume 1: Long Papers)},
  month     = apr,
  year      = {2025},
  address   = {Albuquerque, New Mexico},
  publisher = {Association for Computational Linguistics},
  pages     = {1--23},
  isbn      = {979-8-89176-189-6},
  doi       = {10.18653/v1/2025.naacl-long.1},
  url       = {https://aclanthology.org/2025.naacl-long.1/}
}

@misc{iflytek2021ad,
  author       = {{iFLYTEK}},
  title        = {2021 Advertising Image Classification Competition},
  year         = {2021},
  howpublished = {iFLYTEK AI Developer Competition},
  url          = {https://challenge.xfyun.cn/topic/info?type=ad-2021}
}

\clearpage
\colorlet{red}{RevisionRed}
\appendix

\section{Data Construction and Split Integrity}
\label{app:data}

\subsection{Sources and Cleaning}

The candidate pool consolidates released Chinese and English image--text
pairs from the six resources in Table~\ref{tab:source-provenance}. We retain
only the paired image and content text and re-annotate every instance under a
common schema; inherited task labels, retrieval queries, source identifiers,
and hashtags are excluded from the benchmark inputs and annotation interface.

\begin{table*}[t]
\centering
\small
\setlength{\tabcolsep}{4pt}
\begin{tabular}{@{}p{0.36\textwidth}p{0.32\textwidth}p{0.17\textwidth}@{}}
\toprule
Resource & Domain & Language \\
\midrule
MultiMET~\cite{zhang2021multimet} & Social media & English \\
MultiCMET~\cite{zhang2023multicmet} & Advertisements & Chinese \\
MultiMM~\cite{yang2025multimm} & Mixed & Chinese/English \\
EmoMeta~\cite{lu2025emometa} & Advertisements & Chinese \\
Advertisement Dataset~\cite{ye2021interpreting} & Commercial/public-service
ads & English \\
iFLYTEK 2021 Competition~\cite{iflytek2021ad} & Commercial ads & Chinese \\
\bottomrule
\end{tabular}
\caption{Source composition of the candidate pool.}
\label{tab:source-provenance}
\end{table*}

We remove records with missing, corrupted, or unreadable images; missing or
incomplete paired text; or invalid content that prevents reliable annotation.
Cross-source copies are collapsed before annotation, retaining one canonical
record and its provenance.

\subsection{Duplicate Detection and Challenge Selection}

Exact duplicate detection uses image SHA-256 and NFKC-normalized text. We retain 160
identical normalized texts paired with different images because text reuse
alone does not constitute a duplicate multimodal sample.

For near-duplicate detection, pHash, dHash, and aHash provide a broad image
candidate screen, followed by pixel correlation and normalized mean absolute
error on $256\times256$ grayscale images. Character-level TF--IDF and sequence
similarity screen the paired texts. A candidate is removed only when both
modalities satisfy the near-duplicate rule: image correlation is at least
$0.98$, image error is at most $0.06$, and text similarity is at least $0.90$.
For empty text pairs, image error must be at most $0.05$. We remove one
redundant item from each of nine near-duplicate pairs and discard two
additional unreadable images, leaving unique valid candidates. The
released manifest preserves hashes, similarity scores, and removal decisions.

The five MLLMs then annotate every retained candidate using the complete-tuple
rule in Appendix~\ref{app:protocol}. The 1,000 candidates that fail five-model
agreement undergo human annotation and then form the \MthreeRBench{}. Because challenge
selection is conducted within the shared source mix, \MthreeRBench{} measures
performance on difficult in-domain cases.

\section{Annotation Protocol}
\label{app:protocol}

\subsection{Structured Labels and Consensus}

Each sample is assigned $y=(m,t,s,e)$, where $m$ denotes metaphor occurrence,
$t$ and $s$ denote the most salient Target-Source pair, and $e$ denotes sentiment.
A sample is labeled metaphorical only when the image, text, or their
interaction supports a directional cross-domain mapping for the complete
image--text pair. For such samples, the Target is the concept being described,
whereas the Source supplies the properties projected onto the Target.
Annotators retain the most salient Target-Source pair that best explains the
communicative intent, verify its direction using the paraphrase ``Target is
represented or understood as Source,'' and prefer grounded concrete entities
to unsupported abstract themes. For non-metaphorical samples,
$t=s=\texttt{None}$. Table~\ref{tab:metaphor-rules} summarizes the
operational inclusion and exclusion boundaries.

\begin{figure}[t]
\centering
\includegraphics[width=\columnwidth]{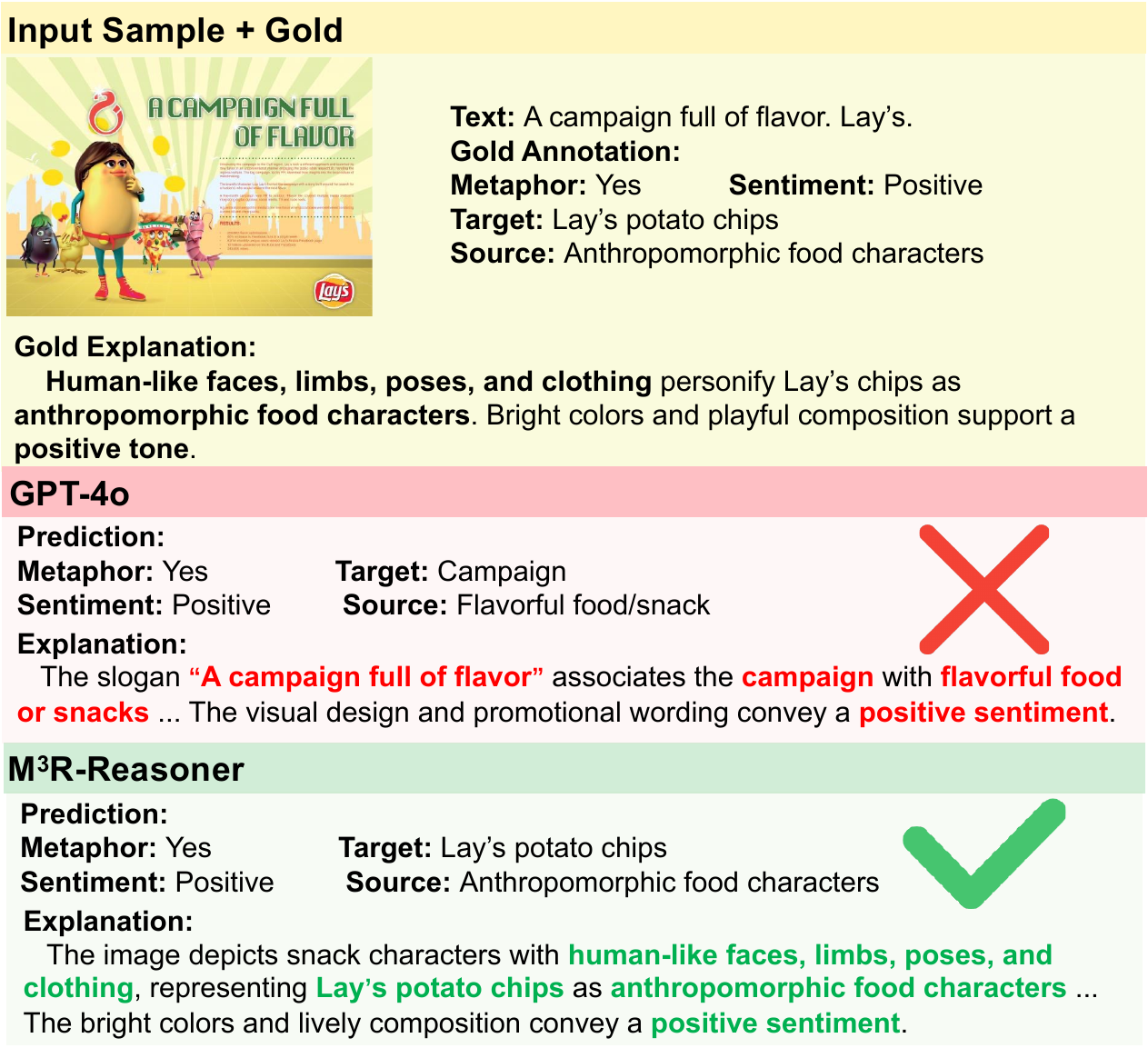}
\caption{Qualitative comparison between GPT-4o and \MthreeRReasoner{}:
text-driven versus visually grounded Target--Source mapping.}
\label{fig:case_study}
\end{figure}

\begin{figure}[t]
\centering
\includegraphics[width=\columnwidth]{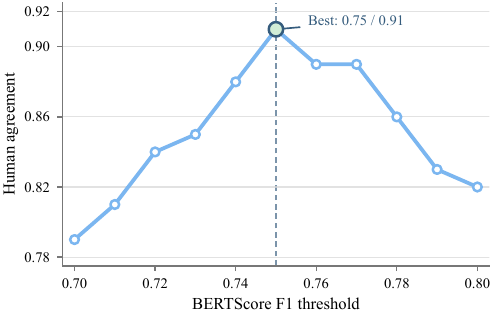}
\caption{BERTScore F1 threshold calibration against human
judgments on 200 instances; the highlighted
point denotes the selected threshold.}
\label{fig:threshold-calibration}
\end{figure}

\begin{table*}[t]
\centering
\small
\setlength{\tabcolsep}{4pt}
\renewcommand{\arraystretch}{1.05}
\begin{tabular}{p{0.15\textwidth}p{0.38\textwidth}p{0.38\textwidth}}
\toprule
Mechanism & Inclusion criterion & Exclusion boundary \\
\midrule
Graphic fusion &
Entities from different semantic domains form one composite. &
Ordinary overlap or co-occurrence without structural or identity fusion. \\
Graphic substitution &
A component of one entity is replaced by an entity from another domain. &
Decoration or occlusion without a cross-domain contribution. \\
Shape borrowing &
Text, a logo, product, or layout adopts the shape of another entity. &
Generic typography or coincidental visual similarity. \\
Personification &
A non-human entity receives human features, behavior, identity, or a social role. &
The ordinary co-presence of a person and a non-human entity. \\
Symbolic transformation &
The Target is represented by a concrete symbol that projects a specific property. &
Theme-level symbolism without an identifiable Target--Source pair. \\
Text-grounded mapping &
The text uses a visible entity to conceptualize another object or state. &
Slogans, exaggeration, or affective wording without a directional entity-level mapping. \\
\bottomrule
\end{tabular}
\caption{Operational inclusion and exclusion criteria for metaphor occurrence.} \label{tab:metaphor-rules}
\end{table*}

Sentiment is assigned from the overall affective stance of the complete
image--text pair: favorable evaluation is positive, criticism or warning is
negative, and primarily descriptive, functional, or sales-oriented content
without a clear affective stance is neutral. For mixed-valence cases,
annotators follow the dominant communicative focus.

\paragraph{Unified Model Instruction.}
All five MLLMs receive the same semantic instruction and four-field JSON
schema; provider-specific wrappers change only the API syntax. The instruction
requires models to apply the boundaries above, verify Target--Source direction,
use the sample's dominant language for the mapping fields, assign sentiment
from the complete pair, and return only
\texttt{MetaphorOccurrence}, \texttt{Target}, \texttt{Source}, and
\texttt{Sentiment}.

Figure~\ref{fig:annotation-prompt} reproduces the complete semantic
instruction supplied independently to all five MLLMs.

\begin{figure*}[p]
\color{black} \centering
\begin{promptbox}[title={Unified Five-Model Annotation
Prompt},coltext=black,colbacktitle=black!10,coltitle=black] \footnotesize
\textbf{Role.} You are an expert in multimodal metaphor and sentiment label
cleaning. Given an image and paired text, determine whether the sample contains
an entity-level metaphor. If so, identify the single most central Target and
Source. Independently assign the overall sentiment.

\medskip
\textbf{Decision rules.} Accept a metaphor only when the image, text, or their
interaction supports a directional cross-domain mapping. Consider graphic
fusion, substitution, shape borrowing, personification, symbolic
transformation, and text-dominant mappings grounded by a visible entity. Reject
ordinary co-occurrence, ordinary displays, slogans without a specific mapping,
theme-only association, and unsupported abstraction. Verify the direction with
``Target is represented or understood as Source.'' Use the sample's dominant
language for Target and Source.

\medskip
\textbf{Sentiment.} Select \texttt{Positive} for an overall favorable attitude,
\texttt{Negative} for an overall unfavorable or warning-oriented attitude, and
\texttt{Neutral} for primarily descriptive, functional, or sales-oriented
content without a clear affective stance.

\medskip
\textbf{Output.} Reason internally and return JSON only:\\
\texttt{\{"MetaphorOccurrence":"Yes/No",}\\
\texttt{"Target":"entity or None",}\\
\texttt{"Source":"entity or None",}\\
\texttt{"Sentiment":"Positive/Neutral/Negative"\}.}

\medskip
\textbf{Input placeholders.}\\
\texttt{Image: \{image\}}\\
\texttt{Text: \{text\}}
\end{promptbox}
\caption{{\color{black}Unified annotation prompt applied independently to the five
MLLMs during candidate labeling.}}
\label{fig:annotation-prompt}
\end{figure*}

Grok-4.1, Kimi-K2.5, GPT-5.5, Claude-Opus-4.7, and Gemini-3.1-Pro independently
produce the complete tuple. A candidate satisfies five-model agreement only
when the same subset of at least three outputs agrees on every applicable
field: exact agreement for metaphor occurrence and sentiment, BERTScore F1
$\geq0.75$ separately for Target and Source when $m=\text{Yes}$, and
\texttt{None} for both mapping fields otherwise. Field-wise majorities from
different model subsets are never combined.

The 1,000 samples without complete-tuple consensus form the Challenge set.
Three trained postgraduate annotators independently label each sample while
viewing the image, text, and anonymized model proposals. Unanimous agreement
on the complete tuple finalizes 693 samples. A senior annotator re-examines
the remaining 307 samples and adjudicates one complete tuple rather than
combining field-wise votes.

\subsection{Annotators, Training, and Compensation}

The three primary annotators were postgraduate students with backgrounds in
computer science or psychology. Before annotation, each completed a two-hour
standardized training session covering entity-level mappings versus thematic
associations, the mechanisms and exclusion boundaries in
Table~\ref{tab:metaphor-rules}, Target--Source direction, concrete versus
abstract granularity, and sentiment from the complete multimodal meaning.
Training includes boundary cases such as ordinary product displays,
promotional exaggeration, image--text thematic agreement without a mapping,
personification, and public-service warnings. The same ordered checklist is
used for independent annotation and adjudication.

Across gold-label annotation and reference-explanation verification, each
primary annotator spent approximately 100 hours and was compensated at
US\$10 per hour (approximately US\$1,000 per annotator). These figures cover
the three primary annotators; senior-adjudication time is not included.

\subsection{Agreement and Threshold Calibration}
\label{app:agreement}

We use two disjoint 200-instance subsets sampled from metaphorical examples
outside the Challenge set. On $\mathcal{D}_{\mathrm{agr}}$, three
annotators independently label all four fields. We report categorical and
open-ended agreement separately because their statistics are not directly
comparable.

\begin{table}[t]
\centering
\small
\begin{tabular}{lcc}
\toprule
Field & Statistic & Value \\
\midrule
Metaphor occurrence & Fleiss' $\kappa$ & 0.85 \\
Sentiment & Fleiss' $\kappa$ & 0.81 \\
Target & Semantic agreement & 0.71 \\
Source & Semantic agreement & 0.68 \\
Complete tuple & Four-field agreement & 0.64 \\
\bottomrule
\end{tabular}
\caption{Inter-annotator agreement on $\mathcal{D}_{\mathrm{agr}}$.}
\label{tab:field-kappa}
\end{table}

On the separate $\mathcal{D}_{\mathrm{cal}}$, humans judge whether predicted
Target and Source concepts are semantically consistent with the gold labels.
Using \texttt{bert-base-multilingual-cased} at layer 9, we evaluate thresholds
from 0.70 to 0.80 in increments of 0.01. As shown in
Figure~\ref{fig:threshold-calibration}, the threshold 0.75 yields the highest
agreement with the binary human judgments (0.91). We therefore use 0.75
consistently for construction-stage semantic matching and JLA binarization.

\section{Reasoning Supervision and Training}

\subsection{Reasoning Corpus}
\label{app:reasoning-corpus}

Five MLLMs independently assign the complete tuple $(m,t,s,e)$ to each
candidate. Candidates proceed to explanation generation only when the same
subset of at least three models agrees on every applicable field under the
rule in Appendix~\ref{app:protocol}. We generate and verify three cumulative
stages for each candidate: $R^{(1)}$ identifies evidence relevant to metaphor
occurrence; $R^{(2)}$ describes the Target--Source mapping and projected
attributes; and $R^{(3)}$ connects the multimodal meaning to sentiment. We
then discard malformed outputs, unsupported inferences, and explanations
inconsistent with their structured labels, leaving 8,989 instances in the
task-specific supervision corpus. The curriculum targets are
\begin{align}
\mathcal{C}^{(1)} &: R^{(1)} \rightarrow m,\\
\mathcal{C}^{(2)} &: R^{(1)}\oplus R^{(2)} \rightarrow (m,t,s),\\
\mathcal{C}^{(3)} &: R^{(1)}\oplus R^{(2)}\oplus R^{(3)}
\rightarrow (m,t,s,e).
\end{align}
For non-metaphorical samples, $R^{(2)}$ is omitted and $t=s=\texttt{None}$.
Each target encloses the cumulative reasoning in \texttt{<think>} tags and the
structured tuple in \texttt{<label>} tags.

\subsection{Training Configuration}
\label{app:training-controls}

All SFT variants use Qwen3-VL-8B with LoRA rank 8 on all target modules, a
4,096-token cutoff length, per-device batch size 1, gradient
accumulation over 8 steps, learning rate $1\times10^{-5}$, cosine decay, a
0.03 warm-up ratio, BF16, and seed 42. Curriculum SFT trains on
$\mathcal{C}^{(1)}\rightarrow\mathcal{C}^{(2)}\rightarrow\mathcal{C}^{(3)}$, with metaphor-to-non-metaphor ratios of
$5{:}5$, $8{:}2$, and $6{:}4$. Each stage initializes its adapter from the
preceding stage. The non-curriculum CoT baseline trains directly on
$\mathcal{C}^{(3)}$ from the base model.

GRPO initializes both policy and reference models from the final curriculum
checkpoint. It uses an actor learning rate of $1\times10^{-6}$, global batch size 40, four responses per prompt, maximum prompt/response
lengths of 1,536/1,024 tokens, and KL coefficient 0.01. The policy uses LoRA
rank 8 with scaling factor 16.

\section{Gold Explanations and Evaluation}

\subsection{Reference Construction}
\label{app:gold-exp}

After the Challenge-set labels are finalized, GPT-5.4 applies the
three stage-specific templates in Figures~\ref{fig:reference-prompt-ve}--\ref{fig:reference-prompt-sj} to predict the complete structured tuple and its
evidence-grounded explanation from the image--text pair. We compare its
predicted tuple with the human gold tuple. A complete label match routes the
generated explanation to two annotators, who independently check its grounding
and minimally revise unsupported or hallucinated content. If any applicable
label mismatches, the annotators instead write the reference explanation from
the original image--text pair and gold labels. A third annotator adjudicates
disagreements in both paths. This process yields 1,000 Image-Text Evidence
references, 510 Mapping Correctness references, and 1,000 Sentiment
Justification references.

The templates are applied in stage order. The mapping stage is used
only for metaphorical instances; for non-metaphorical instances, Target and
Source remain \texttt{None}.

\begin{figure*}[p]
\color{black} \centering
\begin{promptbox}[title={Reference Prompt 1: Image-Text
Evidence},coltext=black,colbacktitle=black!10,coltitle=black] \footnotesize
\textbf{Task.} Determine whether the image--text pair contains an entity-level
multimodal metaphor. Identify the main entities and text; inspect fusion,
substitution, shape borrowing, personification, identity transformation, and
symbolic transformation; and, when the image is structurally ordinary, test
whether the text uses a visible entity to understand another object or state.

\medskip
\textbf{Boundary rule.} Do not infer metaphor from an ordinary display,
co-occurrence, advertising language, emotional adjectives, or thematic
symbolism alone. A positive decision requires a directional entity-level pair.
Cite decisive image and text evidence. Keep the rationale within 150 English
words or an equivalent Chinese length.

\medskip
\textbf{Output.}\\
\texttt{\{"cot1":"evidence-based paragraph",}\\
\texttt{"MetaphorOccurrence":"Yes or No"\}.}

\medskip
\texttt{Image: \{image\}}\\
\texttt{Text: \{text\}}
\end{promptbox}
\caption{{\color{black}Stage-1 prompt for Image-Text Evidence
and metaphor detection.}} \label{fig:reference-prompt-ve}
\end{figure*}

\begin{figure*}[p]
\color{black} \centering
\begin{promptbox}[title={Reference Prompt 2: Mapping
Correctness},coltext=black,colbacktitle=black!10,coltitle=black] \footnotesize
\textbf{Task.} Identify the single most central Target--Source pair. The Target
is the object actually described, understood, or evaluated; the Source is the
entity, form, symbol, identity, function, or attribute borrowed to express it.
Determine whether the trigger is visual-dominant or text-dominant and verify
the direction with ``Target is represented or understood as Source.''

\medskip
\textbf{Constraints.} Prefer observable or clearly referenced entities, avoid
unsupported abstract themes, do not reverse the pair, and choose the mapping
that best explains the image--text intent. Use concise nouns or noun phrases in
the dominant language. Keep the rationale within 150 English words or an
equivalent Chinese length.

\medskip
\textbf{Output.}\\
\texttt{\{"cot2":"direction-and-evidence paragraph",}\\
\texttt{"Target":"target-domain entity",}\\
\texttt{"Source":"source-domain entity"\}.}

\medskip
\texttt{Image: \{image\}}\\
\texttt{Text: \{text\}}
\end{promptbox}
\caption{{\color{black}Stage-2 prompt for Target--Source mapping.}} \label{fig:reference-prompt-mc}
\end{figure*}

\begin{figure*}[p]
\color{black} \centering
\begin{promptbox}[title={Reference Prompt 3: Sentiment
Justification},coltext=black,colbacktitle=black!10,coltitle=black] \footnotesize
\textbf{Task.} Infer the sentiment of the complete image--text pair. Identify
affective cues in the image and text; when a metaphor exists, determine how the
mapping changes or reinforces the value attitude. Do not classify from one word
or local visual element.

\medskip
\textbf{Labels.} Use \texttt{Positive} for praise, hope, care, warmth,
celebration, growth, or protection; \texttt{Negative} for criticism, warning,
danger, harm, disease, fear, oppression, destruction, or loss; and
\texttt{Neutral} for objective information, ordinary display, product or
function description, price promotion, or sales-oriented content without a
clear affective stance. Follow the final semantic focus in mixed cases.

\medskip
\textbf{Output.}\\
\texttt{\{"cot3":"integrated sentiment evidence",}\\
\texttt{"Sentiment":"Positive, Neutral, or Negative"\}.}

\medskip
\texttt{Image: \{image\}}\\
\texttt{Text: \{text\}}
\end{promptbox}
\caption{{\color{black}Stage-3 prompt for Sentiment
Justification.}} \label{fig:reference-prompt-sj}
\end{figure*}

\subsection{Metric Implementation and Edge Cases}
\label{app:evaluation}

MD and SC use all 1,000 instances. TP and SP are conditional
mapping metrics computed over the fixed set $\mathcal{M}$ of $N_M=510$
gold-metaphorical instances. For each $i\in\mathcal{M}$, a correctly detected
metaphor with a valid mapping field receives its Target or Source BERTScore; a
metaphor false negative, parsing failure, or invalid mapping field receives
zero and remains in the denominator. Gold non-metaphors are excluded from
TP/SP regardless of the predicted label because no gold mapping is defined for
them; metaphor false positives are instead penalized by MD and JLA. Thus,
correctly predicted non-metaphors do not receive perfect TP/SP scores; their
mapping fields are not applicable. TP, SP,
and JLA use \texttt{bert-base-multilingual-cased} at layer 9. JLA requires
metaphor occurrence and sentiment to match exactly and requires Target and
Source BERTScore F1 $\geq0.75$ on a gold metaphor; on a gold non-metaphor,
both fields must be \texttt{None}. The calibrated cutoff of 0.75 is used
consistently for construction-stage semantic matching and for converting
open-ended Target and Source predictions into JLA correctness decisions.

Categorical fields are case-normalized after trimming whitespace. Missing,
invalid, or unparsable required fields are incorrect. If a judge request
fails or omits a requested dimension, that dimension receives zero. These
rules retain every sample in its fixed denominator.

For explanation scoring, ITE is zero when metaphor occurrence is incorrect,
MC is zero unless both Target and Source are correct, and SJ is zero when
sentiment is incorrect. Otherwise, the reference-based judge score is retained
and multiplied by 100 for reporting.

\begin{table}[t]
\centering
\small
\setlength{\tabcolsep}{3pt}
\begin{tabularx}{\columnwidth}{@{}p{0.16\columnwidth}p{0.19\columnwidth}X@{}}
\toprule
Metric & Population & Scoring rule \\
\midrule
MD/SC & 1,000 & Accuracy and Macro-F1. \\
TP/SP & 510 & BERTScore P/F1; invalid or missed mappings score zero. \\
JLA & 1,000 & All four fields must be correct on the same instance. \\
ITE & 1,000 & Reference score if MD is correct; otherwise zero. \\
MC & 510 & Reference score if Target and Source are correct; otherwise zero. \\
SJ & 1,000 & Reference score if SC is correct; otherwise zero. \\
\bottomrule
\end{tabularx}
\caption{Evaluation populations and per-instance gating rules.}
\label{tab:metric-denominators}
\end{table}

\subsection{Controlled Prompt Templates} \label{app:prompting}

Figures~\ref{fig:prompt-p1}--\ref{fig:prompt-p3} reproduce the Direct (P1), Definitions (P2), and Evidence CoT
(P3) templates used in the controlled prompting study. The image is
supplied separately, and \texttt{\{ACCOMPANYING\_TEXT\}} is replaced with the paired text. Formatting is
adapted only for
presentation; all instructions and schema fields match the implementation.

\begin{figure*}[p]
\centering
\begin{promptbox}[title={P1: Direct Structured Prompting}]
\footnotesize
\textbf{[System]}\\
You are a rigorous multimodal annotation assistant. Put all reasoning inside
\texttt{<think>...</think>} and put the final answer inside
\texttt{<label>...</label>}.

\medskip
\textbf{[Instruction]}\\
Please judge the overall label of the sample based on the overall
content, expressive theme, and possible metaphorical mapping relationship of
the current image--text sample.

Please place the concise reasoning process inside the
\texttt{<think></think>} tag,then place the final answer inside the
\texttt{<label></label>} tag. To maintain consistency in the label structure,
the final answer needs to include the four fields MetaphorOccurrence, Target,
Source, and Sentiment. If it does not contain a metaphor, please set both
Target and Source to ``None''. The final labels must be a valid JSON object.
MetaphorOccurrence can only be ``Yes'' or ``No''; Sentiment can only be
``Positive'', ``Neutral'', or ``Negative''.

\medskip
\textbf{Output format:}\\
\texttt{<think>Concise reasoning process about whether the metaphor exists,}\\
\texttt{the Target--Source mapping relationship, and sentiment judgment.</think>}\\
\texttt{<label>\{"MetaphorOccurrence":"Yes",}\\
\texttt{"Target":"target entity/tenor",}\\
\texttt{"Source":"source entity/vehicle",}\\
\texttt{"Sentiment":"Negative"\}</label>}\\
\textbf{or}\\
\texttt{<think>Concise reasoning process about the absence of an entity-level}\\
\texttt{metaphor and the overall sentiment judgment.</think>}\\
\texttt{<label>\{"MetaphorOccurrence":"No",}\\
\texttt{"Target":"None","Source":"None",}\\
\texttt{"Sentiment":"Neutral"\}</label>}

\medskip
\textbf{[Input]}\\
\texttt{Text: \{ACCOMPANYING\_TEXT\}}
\end{promptbox}
\caption{P1 direct-prompt template.} \label{fig:prompt-p1}
\end{figure*}

\begin{figure*}[p]
\centering
\begin{promptbox}[title={P2: Task-Definition-Enhanced Prompting}]
\footnotesize
\textbf{Task definitions and decision rules:}
\begin{enumerate}
\item Entity-level multimodal metaphor occurrence: answer Yes only when the
intended interpretation of the image--text pair contains a salient
cross-domain mapping in which one entity or concept is understood,
represented, or evaluated through another. The supporting evidence may
come from the image, the text, or their interaction. Ordinary co-occurrence,
literal depiction, decorative design, or a rhetorical slogan without a
recoverable Target--Source mapping is not sufficient.
\item Target: the most salient entity or concept being described,
represented, understood, or evaluated.
\item Source: the most salient borrowed entity, form, role, or conceptual
domain used to understand or represent the Target. Choose one
Target--Source pair at a consistent semantic granularity and preserve the
direction Target~$\rightarrow$~Source.
\item If MetaphorOccurrence is No, set both Target and Source to None.
\item Sentiment: the overall attitude conveyed by the joint image--text
message toward its central subject, using exactly one of Positive, Neutral,
or Negative.
\end{enumerate}

Apply these definitions directly. Write one concise reasoning block in the
exact form \texttt{<think>...</think>}, followed by exactly one final
\texttt{<label>...</label>} block. The reasoning should explain the metaphor
decision, the Target--Source mapping when applicable, and the sentiment
judgment, but no fixed stage order is required. Do not output Markdown, a code
fence, or any content outside these two blocks.

\medskip
Analyze only the supplied image and its accompanying text.\\
\texttt{Accompanying text: \{ACCOMPANYING\_TEXT\}}

\medskip
Return all four fields using the exact English keys and label values shown
below. Target and Source content may be written in the language of the sample.\\
\textbf{Required final format:}\\
\texttt{<label>\{"MetaphorOccurrence":"Yes|No",}\\
\texttt{"Target":"a concise concept or None",}\\
\texttt{"Source":"a concise concept or None",}\\
\texttt{"Sentiment":"Positive|Neutral|Negative"\}</label>}
\end{promptbox}
\caption{P2 task-definition prompt template.} \label{fig:prompt-p2}
\end{figure*}

\begin{figure*}[p]
\centering
\begin{promptbox}[title={P3: Evidence--Mapping--Sentiment CoT}]
\footnotesize
\textbf{Task definitions and decision rules:}
\begin{enumerate}
\item Entity-level multimodal metaphor occurrence: answer Yes only when the
intended interpretation of the image--text pair contains a salient
cross-domain mapping in which one entity or concept is understood,
represented, or evaluated through another. The supporting evidence may
come from the image, the text, or their interaction. Ordinary co-occurrence,
literal depiction, decorative design, or a rhetorical slogan without a
recoverable Target--Source mapping is not sufficient.
\item Target: the most salient entity or concept being described,
represented, understood, or evaluated.
\item Source: the most salient borrowed entity, form, role, or conceptual
domain used to understand or represent the Target. Choose one
Target--Source pair at a consistent semantic granularity and preserve the
direction Target~$\rightarrow$~Source.
\item If MetaphorOccurrence is No, set both Target and Source to None.
\item Sentiment: the overall attitude conveyed by the joint image--text
message toward its central subject, using exactly one of Positive, Neutral,
or Negative.
\end{enumerate}

\textbf{Reason through the task in the following order:}
\begin{enumerate}
\item Stage 1 -- Evidence: identify the specific visual evidence and textual
evidence relevant to deciding whether a cross-domain metaphor exists.
\item Stage 2 -- Mapping: if a metaphor exists, identify the most salient
Target and Source, explain the mapping direction, and keep both concepts at
a consistent semantic granularity. If no metaphor exists, explain why the
evidence is literal or non-mapping and use None for both fields.
\item Stage 3 -- Sentiment: infer Positive, Neutral, or Negative and justify
it using the joint message.
\end{enumerate}

Write one concise reasoning block in the exact form
\texttt{<think>[Evidence] ... [Mapping] ... [Sentiment] ...</think>}. Then
provide exactly one final \texttt{<label>} block. Do not output Markdown, a
code fence, or any content outside these two blocks.

\medskip
Analyze only the supplied image and its accompanying text.\\
\texttt{Accompanying text: \{ACCOMPANYING\_TEXT\}}

\medskip
Return all four fields using the exact English keys and label values shown
below. Target and Source content may be written in the language of the sample.\\
\textbf{Required final format:}\\
\texttt{<label>\{"MetaphorOccurrence":"Yes|No",}\\
\texttt{"Target":"a concise concept or None",}\\
\texttt{"Source":"a concise concept or None",}\\
\texttt{"Sentiment":"Positive|Neutral|Negative"\}</label>}
\end{promptbox}
\caption{P3 evidence--mapping--sentiment
prompt template.} \label{fig:prompt-p3}
\end{figure*}

\subsection{Human-Gemini Agreement Study}
\label{app:metric-validation}

\paragraph{Sampling and Systems.}
We sample 200 benchmark instances using strata defined by metaphor occurrence,
language, sentiment, source domain, and metaphor construction where
applicable. The same instances are used to evaluate Claude-Sonnet-4.6,
CoT-SFT + Curr., and \MthreeRReasoner{}, yielding 600 anonymized model
outputs. ITE and SJ are evaluated on all sampled instances, whereas MC is
restricted to gold-metaphorical instances. This preserves the task-valid
denominators used by the benchmark.

\paragraph{Human Evaluation Protocol.}
For each applicable stage, annotators view the original image--text pair, gold
labels, human-verified reference explanation, and anonymized model explanation.
They assign one continuous score in $[0,1]$ that jointly reflects correctness,
completeness, specificity, and support from the image, text, or their
interaction. Two annotators score every output independently without access to
model identities or Gemini scores. For each applicable dimension, the human
score is the arithmetic mean of their two final ratings; these averaged scores
are used to compute all means and correlations in
Table~\ref{tab:human-gemini-validation}. We apply the same prerequisite label
gates used by the automatic evaluator: ITE is zero when metaphor occurrence is
incorrect, MC is zero unless both Target and Source are correct,
and SJ is zero when sentiment is incorrect. MC is not defined for gold
non-metaphors.

\paragraph{Agreement with Gemini.}
We compare the averaged human scores with Gemini using Pearson's $r$ and
Spearman's $\rho$, computed separately for each system and dimension. Across systems, the Pearson $r$ ranges are 0.818--0.920 for ITE,
0.747--0.854 for MC, and 0.841--0.897 for SJ; the corresponding Spearman
$\rho$ ranges are 0.783--0.906, 0.747--0.901, and 0.707--0.827. Human and Gemini means differ by at most 4\% of the scoring scale across
all system--dimension pairs, and both evaluators rank the three systems
identically within every dimension. Table~\ref{tab:human-gemini-validation}
reports the full means and correlation coefficients.
\begin{table*}[t]
\centering
\small
\setlength{\tabcolsep}{5pt}
\begin{tabular*}{\textwidth}{@{\extracolsep{\fill}}llccccc@{}}
\toprule
Model & Dimension & $N$ & Human & Gemini & Pearson $r$ & Spearman $\rho$ \\
\midrule
\multirow{3}{*}{Claude-Sonnet-4.6}
& ITE & 200 & 0.496 & 0.510 & 0.899 & 0.906 \\
& MC & 100 & 0.483 & 0.523 & 0.747 & 0.747 \\
& SJ & 200 & 0.525 & 0.545 & 0.895 & 0.827 \\
\addlinespace[1pt]
\multirow{3}{*}{CoT-SFT + Curr.}
& ITE & 200 & 0.556 & 0.586 & 0.818 & 0.783 \\
& MC & 100 & 0.567 & 0.545 & 0.854 & 0.901 \\
& SJ & 200 & 0.564 & 0.571 & 0.841 & 0.707 \\
\addlinespace[1pt]
\multirow{3}{*}{\MthreeRReasoner{}}
& ITE & 200 & 0.578 & 0.597 & 0.920 & 0.825 \\
& MC & 100 & 0.582 & 0.567 & 0.819 & 0.883 \\
& SJ & 200 & 0.605 & 0.618 & 0.897 & 0.777 \\
\bottomrule
\end{tabular*}
\caption{Human--Gemini agreement on 200 stratified instances. Human and Gemini are mean $[0,1]$ scores, where Human denotes the average of two
annotators' ratings per item; $r$ and $\rho$ are item-level Pearson and Spearman correlations.}
\label{tab:human-gemini-validation}
\end{table*}

\subsection{Reference-Based Explanation Judge}
\label{app:judge}

Gemini-3.1-Pro runs at temperature 0 and receives the gold and predicted
labels together with the corresponding human-verified reference and predicted
stage explanations. For each stage that
passes its label gate, the judge returns a score in $[0,1]$ based on semantic
correctness, specificity, completeness, and agreement with the verified
reference. ITE and SJ are averaged over all 1,000 instances, whereas MC is
averaged over the 510 gold metaphors. Scores are multiplied by 100 for
reporting, and Mean is the arithmetic mean of ITE, MC, and SJ. During benchmark construction, Gemini-3.1-Pro serves only as one of five
proposal models, while humans determine the final labels and explanations. No
Gemini-family model is included among the evaluated systems, so the judge never
scores its own or same-family outputs; the observed human--Gemini agreement and this model-family separation reduce the risk of model-family preference.

Table~\ref{tab:human-gemini-validation} shows that Gemini scores closely track image-aware human judgments; based on this agreement, we use it for full-benchmark explanation scoring.
Figure~\ref{fig:judge-prompt} reproduces the complete instruction used
by the reference-based explanation judge.

\begin{figure*}[p]
\color{black} \centering
\begin{promptbox}[title={Reference-Based Explanation Judge
Prompt},coltext=black,colbacktitle=black!10,coltitle=black] \footnotesize
You are a strict evaluator of explanation quality for multimodal metaphor and
sentiment analysis. Score only the requested dimensions. Compare each predicted
rationale with the corresponding gold rationale. A score of 1 means fully
correct, specific, and supported; a score of 0 means wrong, unsupported,
irrelevant, empty, or contradictory. Intermediate scores reflect partial
correctness, completeness, specificity, and evidence support. Do not reward
unsupported facts. Return one continuous score in $[0,1]$ and a brief
justification for each requested dimension.

\medskip
\texttt{Gold labels: \{gold\_labels\}}\\
\texttt{Gold rationales: \{gold\_rationales\}}\\
\texttt{Predicted labels: \{predicted\_labels\}}\\
\texttt{Predicted rationales: \{predicted\_rationales\}}\\
\texttt{Dimensions to score: \{requested\_dimensions\}}
\end{promptbox}
\caption{{\color{black}Prompt used by Gemini-3.1-Pro for reference-based explanation scoring.}}
\label{fig:judge-prompt}
\end{figure*}

\section{Qualitative Analysis}
\label{app:qualitative-analysis}

\paragraph{Case Study.}
Figure~\ref{fig:case_study} illustrates an evidence--mapping mismatch. The
advertisement personifies Lay's potato chips as human-like food characters.
GPT-4o identifies the metaphor and positive sentiment but follows the slogan
\textit{A campaign full of flavor} toward an unsupported mapping. In contrast,
\MthreeRReasoner{} identifies the anthropomorphic visual features, recovers
the intended Target--Source mapping, and connects both modalities to the
sentiment decision. The example shows why label correctness alone cannot
establish evidence-grounded interpretation.

\end{document}